\documentclass{article}

\usepackage{arxiv}

\usepackage{graphicx}%
\usepackage{multirow}%
\usepackage{amsmath,amssymb,amsfonts}%
\usepackage{amsthm}%
\usepackage{mathrsfs}%
\usepackage[title]{appendix}%
\usepackage{xcolor}%
\usepackage{textcomp}%
\usepackage{manyfoot}%
\usepackage{booktabs}%
\usepackage{algorithm}%
\usepackage{algorithmicx}%
\usepackage{algpseudocode}%
\usepackage{listings}%
\usepackage{subcaption}
\usepackage{url}
\usepackage{hyperref}
\usepackage{natbib}

\title{Continuous modeling of manipulative participation for stance analysis: Online discourse on the 2024 Street Animal Law in Türkiye}

\newif\ifuniqueAffiliation
\uniqueAffiliationfalse

\ifuniqueAffiliation 
\else
\usepackage{authblk}

\newbox{\orcid}\sbox{\orcid}{\includegraphics[scale=0.06]{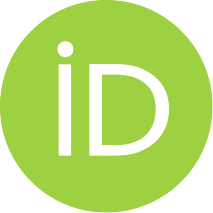}} 

\author[1]{%
    \href{https://orcid.org/0000-0002-8150-4053}{\usebox{\orcid}\hspace{1mm}Ulya Bayram\thanks{Corresponding author: \texttt{ulya.bayram@comu.edu.tr}}}%
}
\author[2]{%
    \href{https://orcid.org/0000-0003-1108-3684}{\usebox{\orcid}\hspace{1mm}Ebru Belkıs Güzeloğlu\thanks{\texttt{ebru.guzeloglu@ege.edu.tr}}}%
}
\author[3]{%
    \href{https://orcid.org/0000-0001-7232-9233}{\usebox{\orcid}\hspace{1mm}Emel Kuşku Özdemir\thanks{\texttt{emel.kusku.ozdemir@ege.edu.tr}}}%
}
\author[4]{%
    \href{https://orcid.org/0000-0001-5111-927X}{\usebox{\orcid}\hspace{1mm}Elif Üstündağlı Erten\thanks{\texttt{elif.ustundagli@ege.edu.tr}}}%
}
\author[5]{%
    \href{https://orcid.org/0000-0001-7805-3979}{\usebox{\orcid}\hspace{1mm}Öyküm Esra Yiğit\thanks{\texttt{oeyigit@yildiz.edu.tr}}}%
}

\affil[1]{Department of Electrical and Electronics Engineering, Faculty of Engineering, Çanakkale Onsekiz Mart University, Çanakkale, 17110, Türkiye}
\affil[2]{Department of Public Relations and Publicity, Faculty of Communication, Ege University, Izmir, 35100, Türkiye}
\affil[3]{Department of Media and Communication, Faculty of Communication, Ege University, Izmir, 35100, Türkiye}
\affil[4]{Department of Business Administration, Faculty of Economics and Administrative Sciences, Ege University, Izmir, 35100, Türkiye}
\affil[5]{Department of Statistics, Yildiz Technical University, Istanbul, 34349, Türkiye}
\fi

\renewcommand{\undertitle}{This work has been submitted to Journal of Computational Social Science for possible publication.}

\hypersetup{
pdftitle={Continuous modeling of manipulative participation for stance analysis: Online discourse on the 2024 Street Animal Law in Türkiye},
pdfsubject={q-cs.AI},
pdfauthor={Ulya Bayram, Ebru Belkıs Güzeloğlu, Emel Kuşku Özdemir, Elif Üstündağlı Erten, Öyküm Esra Yiğit},
pdfkeywords={Computational social science; Manipulative participation; Stance analysis; Political discourse; Social media; Natural language processing},
}

\begin{document}
\maketitle

\begin{abstract}
	Studies of online political discourse commonly infer public opinion through stance analysis while implicitly assuming that observed user participation is authentic. Meanwhile, studies focusing on detecting manipulative participation are typically modeling it as a binary phenomenon, overlooking the varying degrees to which users may engage in coordinated or inauthentic behavior. This study proposes a continuous modeling approach for manipulative participation and examines its integration with stance analysis. Using online discourse on the 2024 Street Animal Law in Türkiye as a case study, we collected a dataset comprising 7,320,574 tweets from 537,994 users. From this corpus, 301 users were selected through a high-information sampling strategy and manually annotated by domain experts using a 10-criterion rubric that yields user-level continuous manipulative participation scores. Evaluation under a unified user-level cross-validation protocol showed that the proposed framework accurately estimates both manipulative participation and user stance, with gradient boosting methods providing the best manipulative participation estimates and BERTurk achieving the highest stance classification performance. External validation on an independently annotated user set demonstrated good generalization performance. The findings support modeling manipulative participation as a continuous construct rather than a binary label, providing a more informative basis for computational analyses of the relationship between manipulative participation and stance in online political discourse.
\end{abstract}


\section{Introduction}

Social media has become an increasingly important source for studying public opinion on controversial political and policy issues. Computational analyses of online discourse, particularly stance analysis, are frequently used to examine public attitudes toward emerging events and legislative proposals. However, these analyses often implicitly assume that observed online participation reflects authentic user behavior. During highly polarized debates, this assumption may not hold, as coordinated or otherwise manipulative participation can disproportionately influence the visibility and distribution of opinions. Consequently, inferences drawn solely from observed stance distributions may provide a distorted picture of public opinion \citep{reveilhac2022systematic}.

This study examines one such case in Türkiye: the 2024 Street Animal Law debate. The Animal Protection Law (No. 5199), enacted in 2004, established welfare obligations for owned and stray animals, including vaccination, neutering, and sheltering. However, inconsistent implementation over time transformed the issue into an increasingly polarized political debate involving competing advocacy coalitions emphasizing either animal welfare or public safety. Following the emergence of organized advocacy campaigns in 2022, discussion intensified on social media platforms such as X, where coordinated messaging and large-scale hashtag campaigns shaped the visibility of competing narratives. Yet, despite the intensity of online discourse, survey evidence collected from the real public suggested that the issue was not among their highest priorities and that support for extreme positions were actually limited \citep{GercekGundem2024, GazeteOksijen2024}. This discrepancy raises important questions regarding the extent to which observed social media discourse reflects authentic public opinion.

Addressing this challenge requires computational methods capable of distinguishing authentic from manipulative participation while simultaneously estimating user stance. Existing stance analysis research primarily relies on datasets collected and annotated  for natural language processing (NLP) tasks. Such datasets are generally designed to improve predictive performance rather than to examine the authenticity of real-life participation, often retaining both authentic and manipulative accounts. Studies that explicitly address manipulation typically rely on bot detection systems or binary classifiers \citep{yang2022botometer,aldayel2022characterizing,abreu2020twitter,rodriguez2020one}. However, these approaches are largely developed for English-language data, exhibit reduced effectiveness in other languages \citep{rauchfleisch2020false}, and primarily focus on automated accounts while overlooking coordinated human participation and other forms of inauthentic behavior \citep{schoch2022coordination}.

In this study, we use the term \textit{manipulative participation} to broadly describe behaviors intended to artificially influence online discourse through automation, coordination, or other inauthentic practices. Rather than treating manipulative participation as a binary phenomenon, we model it as a continuous construct that captures varying degrees of manipulative behavior. Binary approaches discard variation in behavioral authenticity, limiting analyses of how differing degrees of manipulative participation relate to political stance. Building on this conceptualization, we develop a computational framework that estimates manipulative participation and user stance from Turkish social media discourse. This enables the examination of stance distributions across different levels of manipulative participation and provides a more informative basis for interpreting online political discourse.

We evaluate the proposed approach using a large-scale dataset comprising 7.3 million tweets related to the 2024 Street Animal Law debate in Türkiye. From this corpus, users were selected through a high-information sampling strategy and manually annotated by domain experts using a literature-grounded rubric from which continuous manipulative participation scores were derived. We then systematically compare  classical machine learning methods, multilingual and Turkish-specific transformers, followed by external validation on independently annotated unseen users.

The main contributions of this study are as follows:

\begin{enumerate} 
\item We operationalize manipulative participation as a continuous construct through a literature-grounded expert annotation framework based on ten evaluation criteria.

\item We develop a unified computational framework for estimating manipulative participation and user stance, enabling analyses of how stance distributions vary across different levels of manipulative participation.

\item We provide a comprehensive empirical evaluation of classical machine learning methods and transformers under a unified user-level evaluation protocol, complemented by external validation on independently annotated users.

\item We demonstrate how continuous modeling of manipulative participation facilitates more nuanced analyses of the relationship between manipulative participation and stance in online political discourse.
\end{enumerate}

\section{Related work}

\subsection{Stance detection in online political discourse}

Stance detection aims to classify the attitude expressed in a text toward a target as supporting, opposing, or neutral \citep{kuccuk2020stance}. Recent studies show that incorporating contextual signals beyond the target text, including user history, interaction networks, and prior posts, can improve classification performance \citep{kreutz2022detecting, samih2021few}. These findings indicate that stance cannot always be inferred reliably from isolated texts and benefits from broader contextual information.

Most contemporary stance detection research focuses on improving prediction performance using public social media data \citep{barberia2026sentiment}. However, social media datasets generally do not distinguish between authentic and manipulative participation. Consequently, existing stance analyses implicitly assume that observed participation reflects genuine opinion expression, limiting their ability to distinguish authentic public sentiment from discourse influenced by coordinated or otherwise manipulative behavior.

\subsection{Manipulative participation and information integrity}

Literature focusing on detecting manipulative participation in social media spans bot detection, troll detection, and coordinated behavior analysis. Most studies formulate the problem as a classification task, distinguishing automated or suspicious accounts from genuine users using features derived from account metadata, user behavior, textual content, and network structure \citep{abreu2020twitter, rodriguez2020one}. These approaches have been applied across datasets ranging from thousands to millions of accounts in supervised learning settings.

Bot detection has received particular attention due to the availability of annotated datasets and widely used tools such as Botometer, which applies machine learning classifiers trained on labeled bot corpora using behavioral and linguistic features \citep{yang2022botometer}. Such tools have enabled large-scale empirical analyses of online discourse. For example, a study analyzed over 4,000 users across politically contentious topics and examined their exposure to bot-like content in large-scale interaction networks \citep{aldayel2022characterizing}.

Despite the availability of tools such as Botometer, bot detection systems are sensitive to domain shift and language variation, and their performance may degrade outside the conditions under which they are trained \citep{rauchfleisch2020false}. Furthermore, manipulative participation extends beyond automated accounts. Political influence campaigns frequently involve coordinated human activity, strategic amplification, and other forms of inauthentic behavior that cannot be adequately represented through a simple bot-versus-human distinction \citep{schoch2022coordination, antonakaki2026detecting}.

Troll detection similarly relies on behavioral patterns and contextual information rather than automation alone \citep{wu2019misinformation}. Consequently, such approaches often depend on relatively small annotated datasets and domain-specific definitions of manipulation \citep{molenda2022makes}. While these models can perform well within specific settings, their generalizability across domains, languages, and political contexts remains limited.

A common characteristic of the literature is that manipulative participation is primarily modeled as a binary classification problem and subsequently used as a preprocessing or filtering step. While effective for identifying suspicious accounts, this practice provides limited insight into how varying degrees of manipulative participation relate to observed discourse and stance distributions.

\subsection{Research gap}

Prior research has substantially advanced both stance detection and manipulative participation detection. However, these research areas have largely evolved independently. Stance detection primarily seeks to improve predictive performance while implicitly assuming authentic participation, whereas manipulative participation detection focuses on identifying suspicious accounts that are subsequently filtered from downstream analyses.

This separation limits computational investigations of whether observed stance distributions reflect authentic public opinion or discourse disproportionately shaped by manipulative participation. Moreover, manipulative participation is commonly operationalized as a binary phenomenon despite often exhibiting varying degrees of coordination, amplification, and strategic engagement. Modeling manipulation as a continuous construct offers the opportunity to preserve this variation and investigate its relationship with stance in greater detail.

These limitations are particularly important in multilingual settings, where approaches developed for English-language social media frequently experience performance degradation because linguistic conventions, user behavior, and political discourse differ across languages and socio-political contexts \citep{elmas2023analyzing}. Studies on Turkish social media similarly demonstrate that language-specific models outperform general multilingual approaches for stance-related tasks \citep{alav2025turkish}, highlighting the importance of context-sensitive modeling.

Motivated by these limitations, this study models manipulative participation as a continuous construct and provides a computational framework for estimating manipulative participation and user stance in Turkish social media discourse. This enables the examination of how stance distributions vary across different levels of manipulative participation while remaining adaptable to non-English political discourse.

\section{Methodology}

First, we obtained the raw collection of Turkish X posts related to the street animals debate based on hashtag usage and preprocessed it by removing duplicates and erroneous data. Next, we conducted an exploratory analysis of the collection before applying a high-information sampling strategy for selecting users for annotation. Three expert annotators developed a literature-grounded annotation framework consisting of ten evaluation criteria, which was used to assign continuous manipulative participation scores to the selected users. They also assigned stance labels to each user based on their position in the debate. Using these annotations, we evaluated multiple methods for estimating manipulative participation and classifying user stance. The following subsections describe the research workflow, including the data collection, exploratory analysis, user sampling, annotation protocol, computational framework, and evaluated models.

\subsection{Data collection and analysis}

Our study focuses on X discourse that occurred before and during the legislative process of Law No. 7527 on street animals, which became law on 31 July 2024 in Türkiye. We obtained the data from a national media monitoring company (MTM) in May 2025 as part of an ongoing research project. To capture different phases of the debate, our data collection starts from 1 December 2023, which is the date of the President’s announcement initiating the legislative process, and ends on 5 August 2024, which is shortly after the parliamentary approval of the law. Since manually identifying all relevant users is infeasible, we first conducted an expert review of discussions on X to identify hashtags frequently used in the debate. The selected hashtags translate from Turkish into English as follows: street, animal, stray, ownerless, dog, sterilization, euthanasia.  This selection was further validated by analyzing Google Trends peaks of these hashtags during the selected period (1 December 2023 to 5 August 2024). MTM then identified all users that used at least two of these hashtags within the selected temporal window and collected all of their tweet and retweet activity from this window. The raw collection contains 7,354,446 tweets and retweets. We identified 33,860 duplicates caused by API-level collection artifacts, and 12 null entries. Removing them resulted in a final dataset of 537,994 unique users, and 7,320,574 tweets and retweets, where 5,656,012 (77\%) of them are retweets.

\begin{figure*}[t]
    \centering

    \begin{subfigure}[b]{0.495\textwidth}
        \centering
        \includegraphics[width=\textwidth]{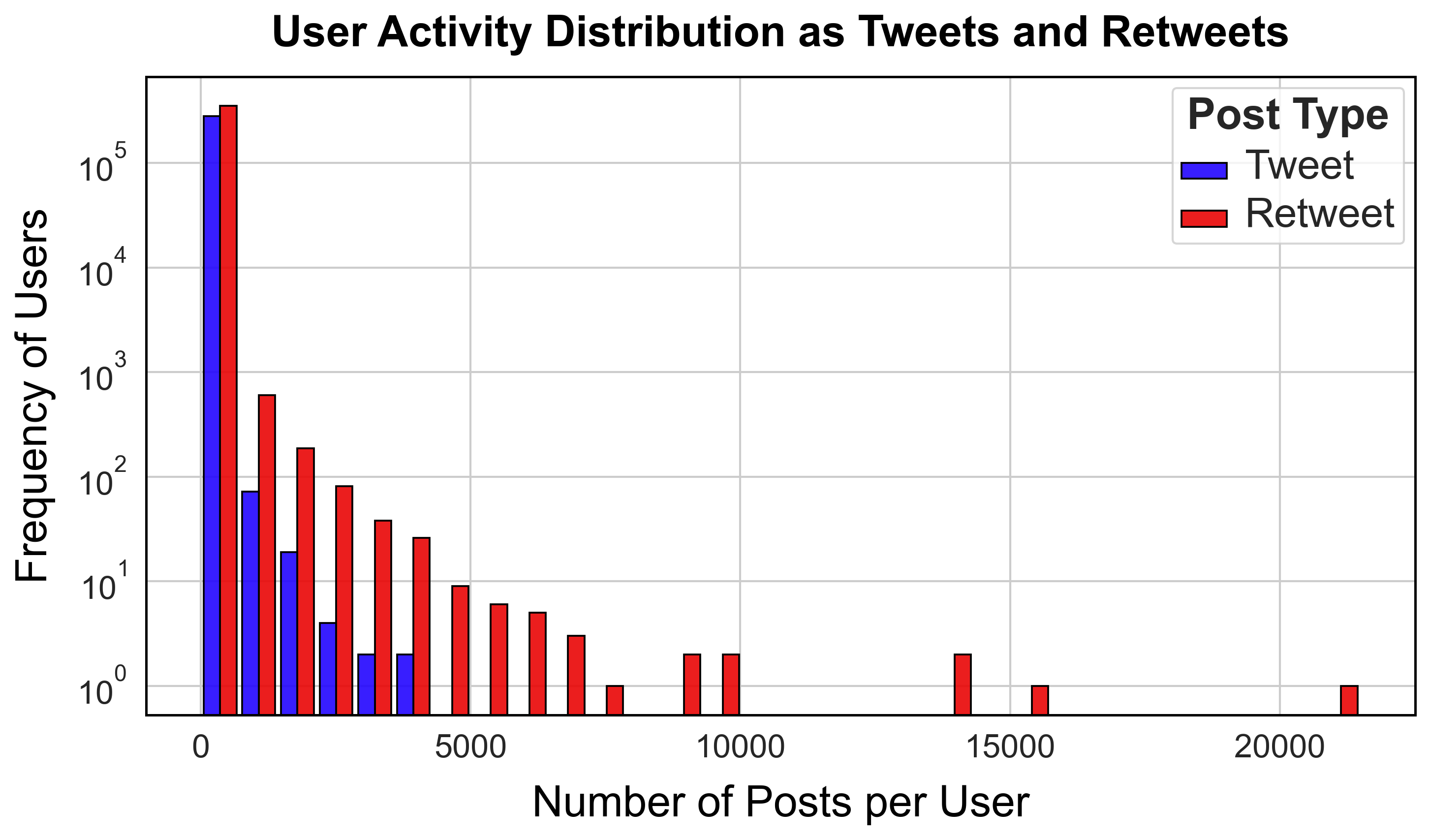}
        \caption{Distribution of tweets and retweets per user.}
        \label{fig:dist_t_vs_rt_per_user}
    \end{subfigure}
    \hfill
    \begin{subfigure}[b]{0.495\textwidth}
        \centering
        \includegraphics[width=\textwidth]{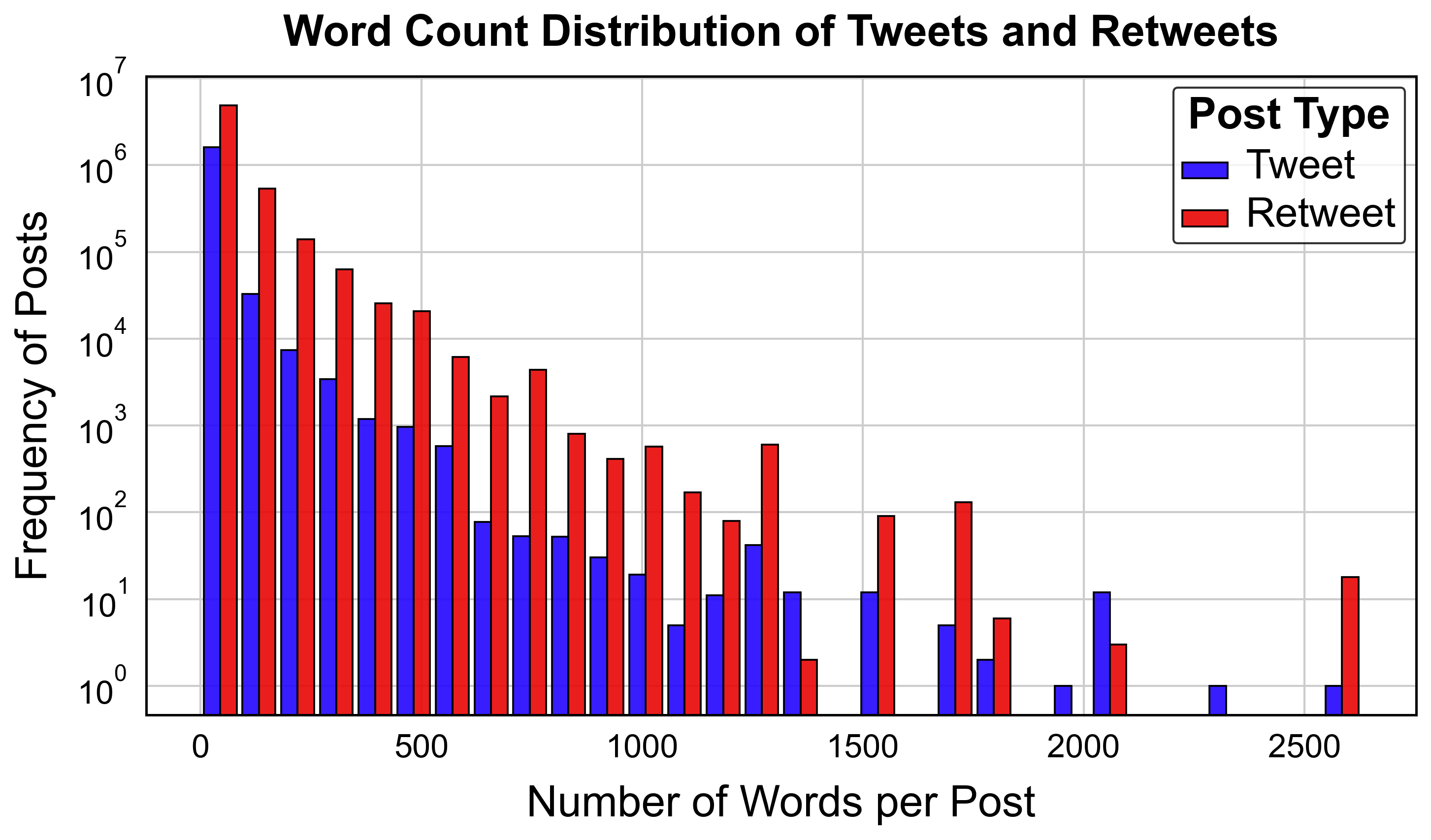}
        \caption{Distribution of tweet and retweet lengths.}
        \label{fig:tweet_len_dist}
    \end{subfigure}
    
    \begin{subfigure}[b]{0.495\textwidth}
        \centering
        \includegraphics[width=\textwidth]{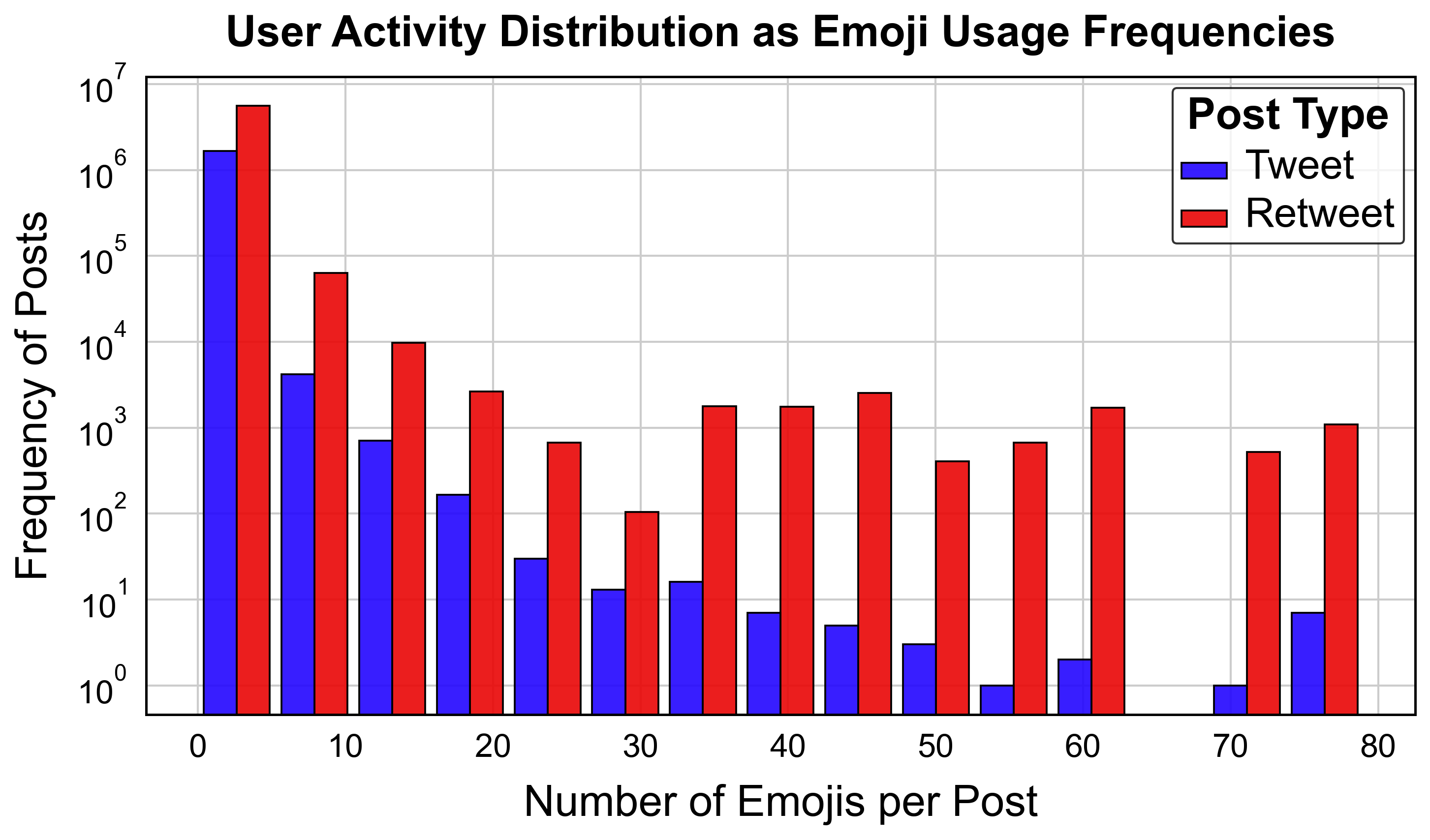}
        \caption{Emoji usage frequencies.}
        \label{fig:emoji_dist}
    \end{subfigure}
    \hfill
    \begin{subfigure}[b]{0.495\textwidth}
        \centering
        \includegraphics[width=\textwidth]{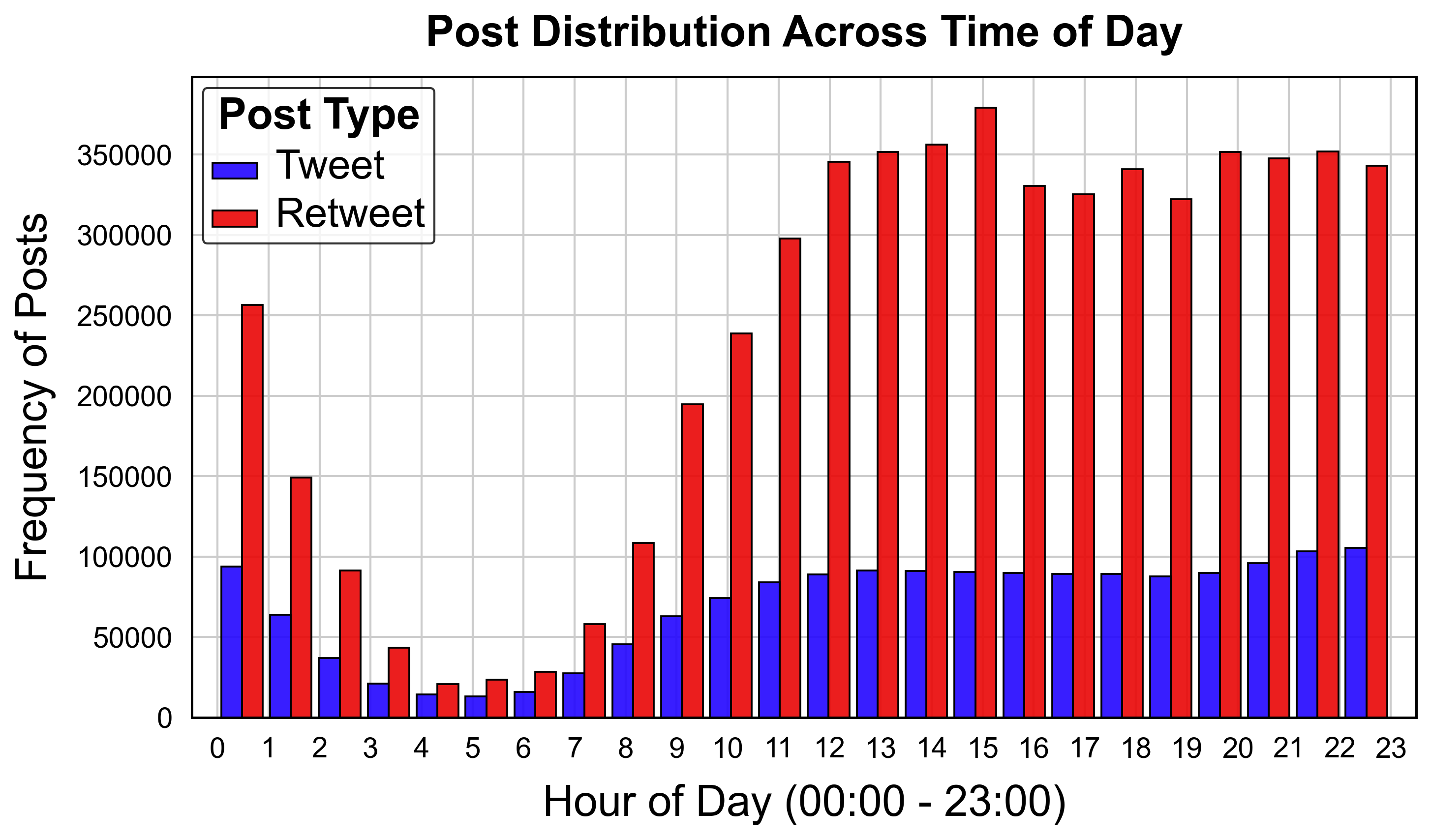}
        \caption{Post distribution across time of day.}
        \label{fig:time_dist}
    \end{subfigure}

    \begin{subfigure}[b]{0.495\textwidth}
        \centering
        \includegraphics[width=\textwidth]{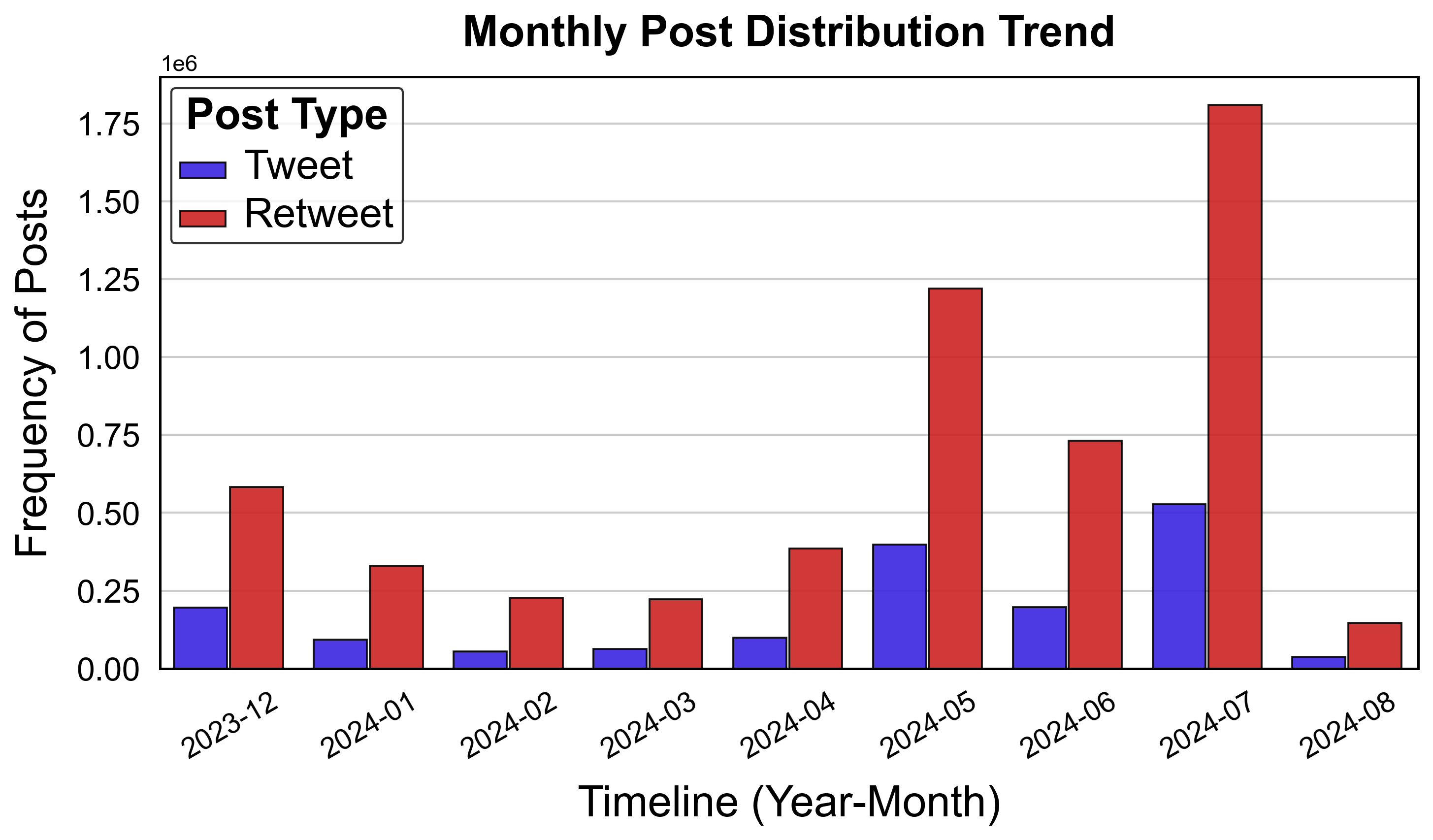}
        \caption{Monthly post distribution.}
        \label{fig:month_dist}
    \end{subfigure}
    \hfill
    \begin{subfigure}[b]{0.495\textwidth}
        \centering
        \includegraphics[width=\textwidth]{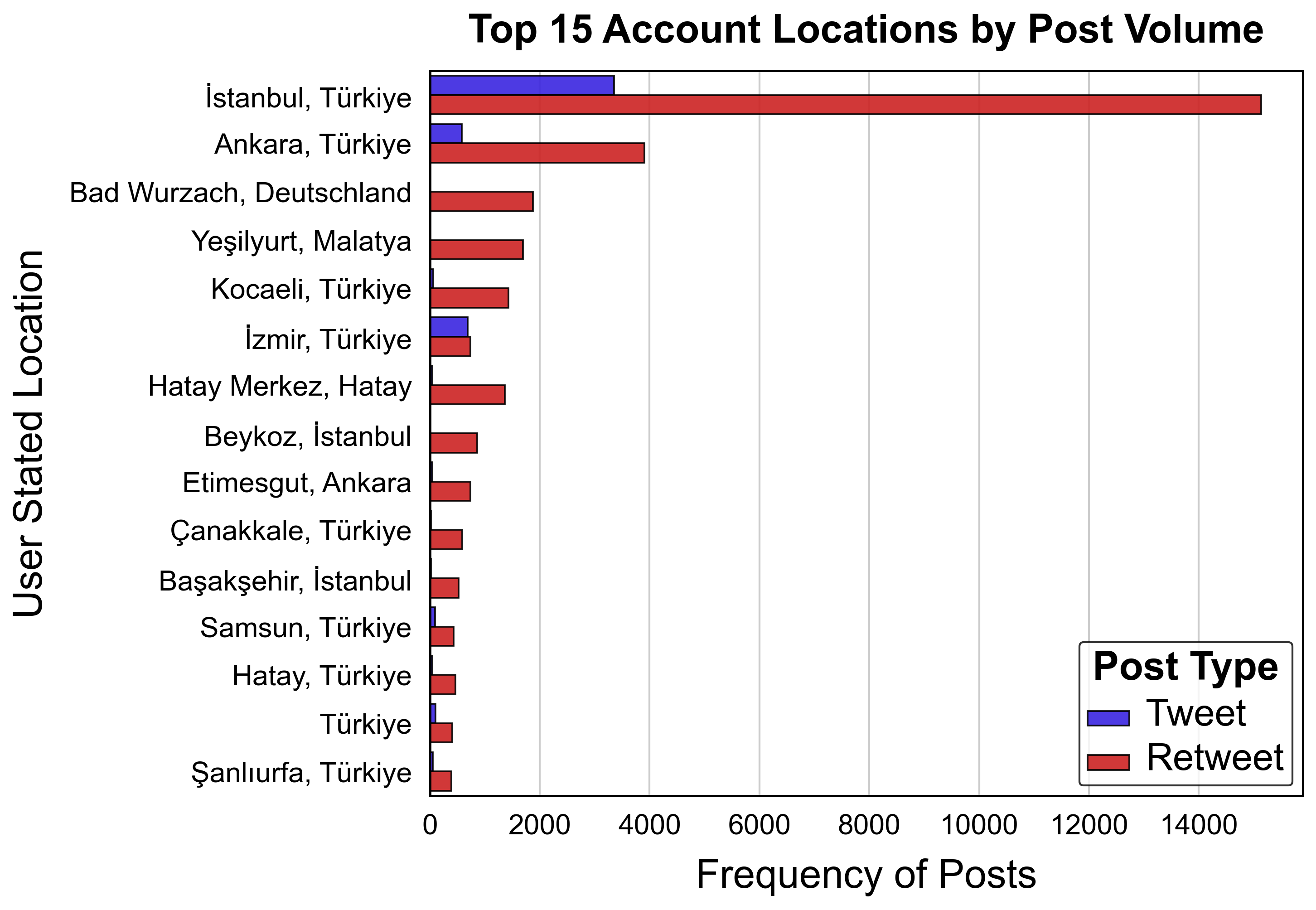}
        \caption{Top 15 locations by post counts.}
        \label{fig:city_dist}
    \end{subfigure}

    \caption{Various statistical distributions within the data collection.}
    \label{fig:stat_plots1}
\end{figure*}


To characterize the underlying patterns and potential anomalies within the collection, we first visualized the distributions of several dataset characteristics. This exploratory analysis provides an overview of user activity patterns and helps identify characteristics relevant to the subsequent annotation and modeling stages. First, we visualized the distribution of the number of tweets and retweets posted by users in Figure~\ref{fig:dist_t_vs_rt_per_user}, which shows no user contributed more than 5,000 original tweets, whereas many accounts produced more than 5,000 retweets. Next, we analyzed the distribution of tweet and retweet lengths, measured as the number of whitespace-separated tokens. Figure~\ref{fig:tweet_len_dist} shows that retweets tend to be longer than original tweets, likely reflecting quote-retweets, which are included in the retweet category by MTM and often contain additional user-generated text. The right tail of the distribution also contains a small number of exceptionally long posts. Post lengths approaching or exceeding 1,000 whitespace-separated tokens are consistent with Premium X accounts, which have supported substantially longer posts since 2023. Finally, emoji usage statistics (Figure~\ref{fig:emoji_dist}) show that retweets contain more emojis on average than original tweets.

We also analyzed the temporal distributions. Figure~\ref{fig:time_dist} shows that posting activity typically begins increasing in the morning, remains high throughout the day, and declines after midnight, reaching its minimum during the early morning hours. This pattern is broadly consistent with typical daily human activity cycles. Figure~\ref{fig:month_dist} further reveals temporal variation associated with the legislative process and the intensity of public discussion. February 2024 exhibits the lowest posting volume in the dataset, followed by March, after which activity increased in April and peaked in July 2024. Although August 2024 appears among the months with the lowest activity levels, the collection period ended midway through the month.

Finally, we examined the geographical distribution of activity. Figure~\ref{fig:city_dist}, which reports the top 15 self-reported user locations by posting volume, shows that large cities such as Istanbul, Ankara, and Izmir account for most original tweets. In contrast, the retweet distribution exhibits notable anomalies. For example, despite contributing relatively few original tweets, Deutschland and Malatya exhibit disproportionately high retweet volumes.

\subsection{High-information sampling}

\begin{figure*}[ht!]
    \centering

    \begin{subfigure}[b]{0.495\textwidth}
        \centering
        \includegraphics[width=\textwidth]{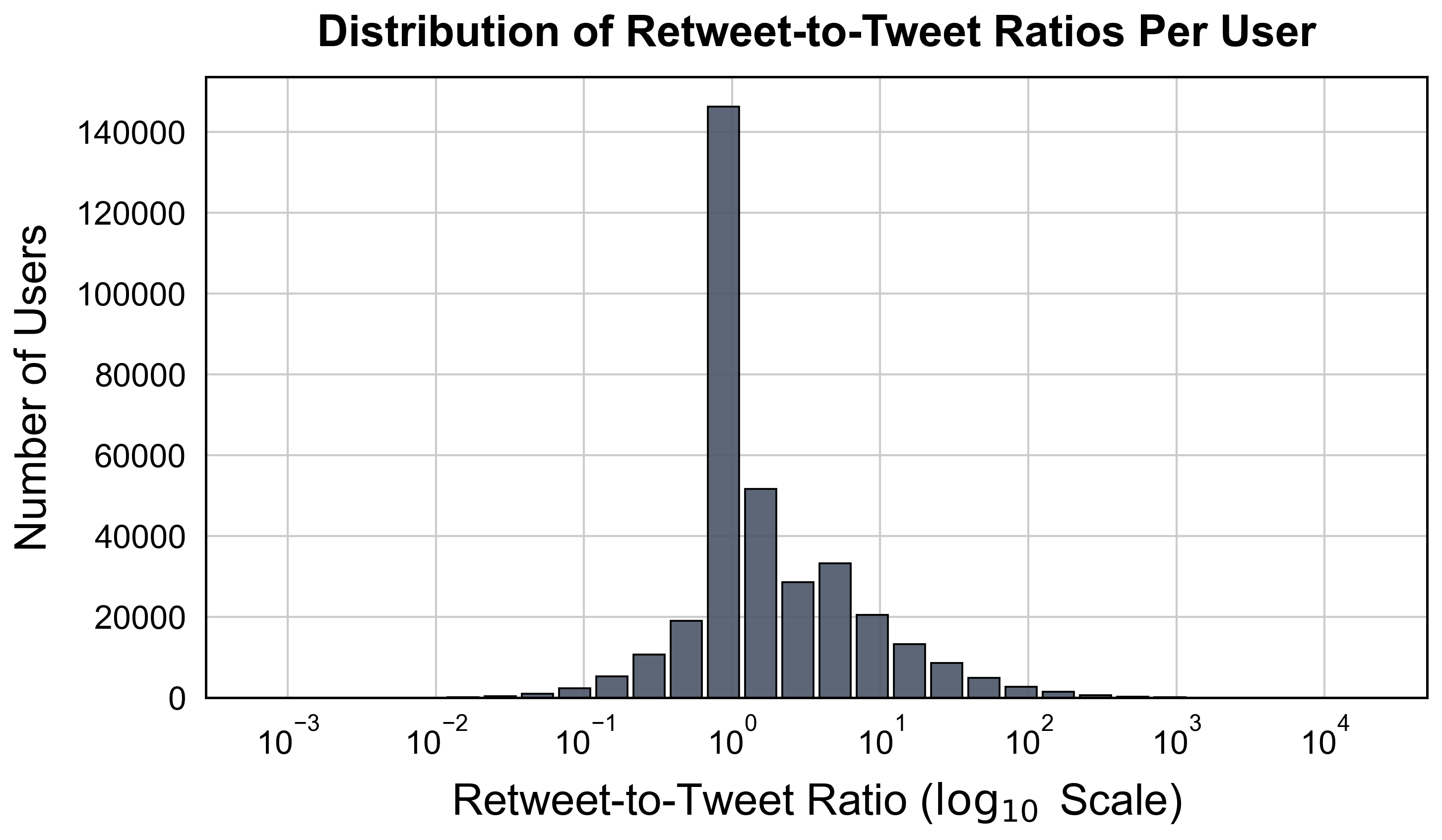}
        \caption{Distribution of retweet to tweet ratios per user.}
        \label{fig:rule_rt_t}
    \end{subfigure}
    \hfill
    \begin{subfigure}[b]{0.495\textwidth}
        \centering
        \includegraphics[width=\textwidth]{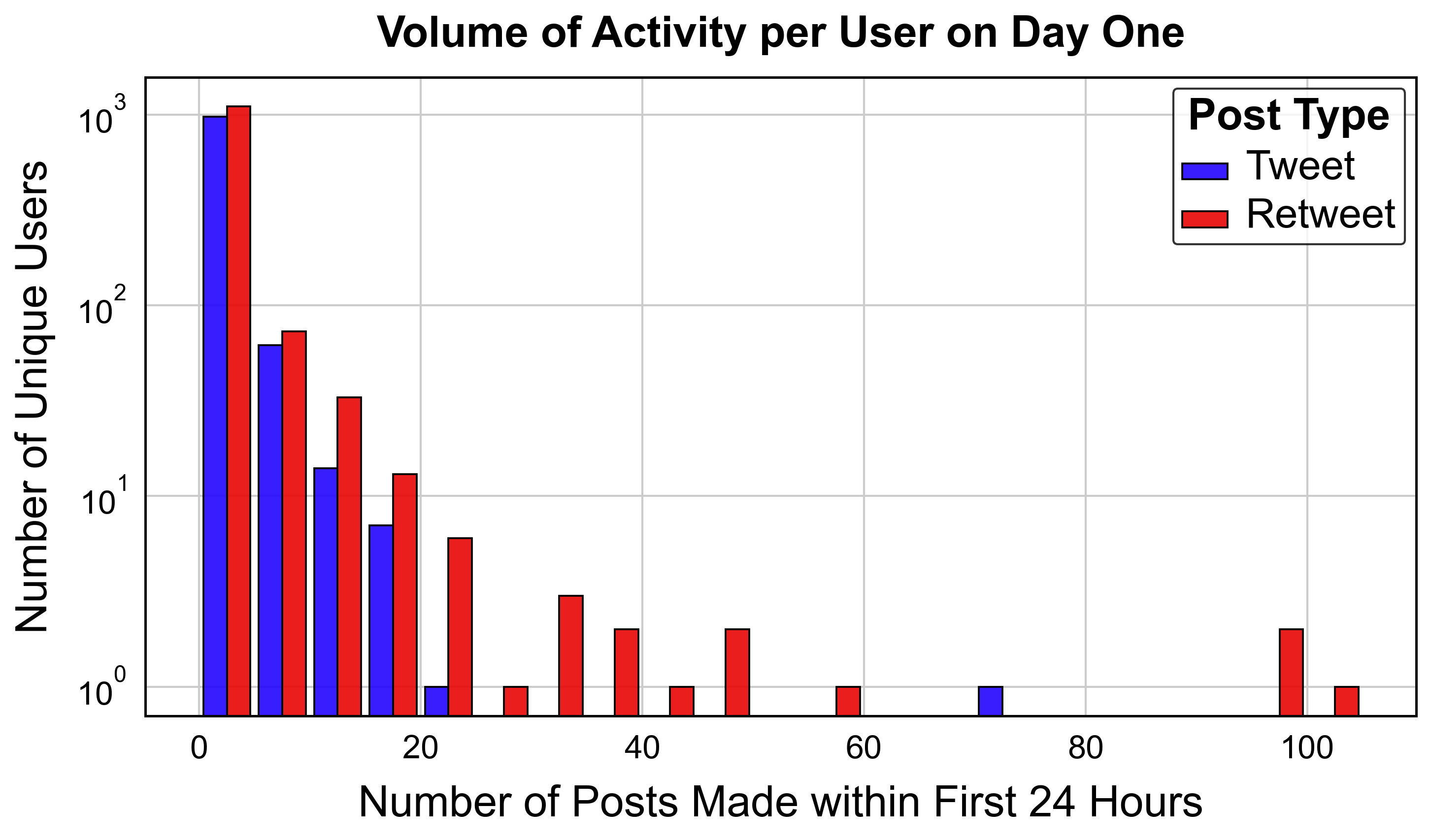}
        \caption{Posting within 24 hours of account creation.}
        \label{fig:rule_24h}
    \end{subfigure}

    \caption{Distributions underlying the high-information sampling strategy.}
    \label{fig:rule_based_sampling_indicators}
\end{figure*}

A recent computational social science study demonstrated that informed sampling strategies can improve the efficiency of expert annotation by prioritizing informative instances over randomly selected ones \citep{zhao2026leveraging}. Building on the exploratory analyses presented in the previous subsection, we adopted a high-information sampling strategy instead of random sampling. 
These heuristics were used solely to prioritize potentially informative accounts for expert review and did not determine the final manipulative participation scores, which were assigned independently using the annotation rubric.

First, we examined the distribution of users' retweet-to-tweet ratios. Figure~\ref{fig:rule_rt_t} shows that most users primarily produce original content while retweeting fewer or a comparable number of posts. In contrast, the long right tail corresponds to users who predominantly engage in retweeting with relatively little original content creation. Such accounts may exhibit unusually strong content amplification and therefore constitute informative candidates for expert annotation. Second, account creation dates were extracted from user metadata and used to calculate the number of tweets and retweets posted during the first 24 hours after account creation. Figure~\ref{fig:rule_24h} shows that most users exhibit limited activity during this period, whereas a small number of accounts display exceptionally high levels of posting, particularly through retweets. These extreme cases represent behavioral outliers and therefore constitute informative candidates for expert annotation.

Based on these exploratory analyses, three groups of accounts were selected for expert annotation. Although the target was to select approximately 100 users from each sampling strategy, a small number of additional users were included to compensate for accounts that might become unavailable or deleted during the expert-annotation process. The first group consisted of the 102 users selected from the extreme right tail of the first-day activity distribution in Figure~\ref{fig:rule_24h}. The second group consisted of the 101 users in the extreme right tail of the retweet-to-tweet ratio distribution in Figure~\ref{fig:rule_rt_t}. Finally, 115 users were selected through random sampling from the complete collection to ensure that the annotation set also included typical users, thereby reducing selection bias while preserving informative cases. Although 318 users may appear small relative to the full collection, manual annotation required three experts to examine more than 150,000 tweets and retweets of these users, which is a time-consuming process.

\subsection{Expert annotation}

Three computational social scientists participated in the expert annotation process. Because manipulative participation cannot be directly observed and requires interpretation of multiple behavioral signals, the annotators first developed a literature-grounded annotation rubric to promote consistency across evaluations. The rubric consists of ten criteria, each independently assessed for every user. Annotators assigned a score of 1 when evidence supporting a criterion was present and -1 for its absence. When a user's account had been deleted before annotation or if there is insufficient evidence to assess a criterion, a score of 0 was assigned. The ten criteria are as follows:

\begin{enumerate}
\item General: Evaluates whether the profile information, including the profile picture, display name, username, and biography, appears to be fake, misleading, incomplete, or non-personal. Also examines the presence of irrelevant content in the profile description, such as references to pornography or sexually suggestive content \citep{suler2004online, coles2016trolling}.

\item Language Use: Assesses linguistic simplicity and repetitiveness in an account’s tweets, checking whether the tweets consist of short, shallow, and frequently repeated expressions, indicating limited semantic depth, sparse use of nouns, pronouns, and adverbs. In addition, examined whether the textual content closely resembles that of other suspected troll accounts, such as the use of highly similar or synonymous phrases across different users, minimal variation in word choice, and limited paraphrasing. These patterns are indicative of coordinated or automated content production, as reported in prior research on troll and bot behavior \citep{tomaiuolo2020survey}.

\item Aggressive Content: Evaluates whether an account’s tweets contain aggressive or hostile language, checking the presence of hate speech, derogatory expressions, or deliberately provocative content. This content can include racist, sexist, homophobic, or misogynistic language that targets individuals, groups or communities based on protected characteristics. This criterion has been used as a feature of troll and coordinated manipulative activities in the literature \citep{chun2019detecting, coles2016trolling}.

\item Topical Coherence: Examines the consistency of narratives expressed in an account’s tweets, assessing whether the tweets convey multiple extreme, unrelated, or mutually contradictory viewpoints over time, if there are abrupt shifts in stance, endorsement of conflicting ideological positions, or the simultaneous promotion of divergent narratives. Such inconsistencies could indicate troll behavior \citep{monakhov2020understanding}.

\item Following/Follower Ratio: Assesses the imbalance between the number of accounts an account follows and the number of followers it has. Such asymmetries are commonly associated with inauthentic or coordinated accounts that seek attention rather than organic social engagement, as noted in prior studies \citep{fornacciari2018holistic}, and used in the present rule-based suspicious account selection process.

\item Coordination: Evaluates the extent to which an account engages in coordinated activity with other accounts, such as frequently sharing identical or highly similar content, and whether it repeatedly retweets or mentions the same set of accounts. Recurrent interaction patterns, synchronized sharing behavior, or persistent engagement within a tightly connected group may indicate troll behavior \citep{cinelli2022coordinated}.

\item Emotional Polarity: Assesses the emotional extremity of an account’s tweets, evaluating whether the content consistently exhibits exaggerated emotional expressions, such as excessive praise or slander, or high-arousal negative emotions, especially anger and rage, all of which can be troll behavior \citep{coles2016trolling, seah2015troll}.

\item Politics: Assesses whether there is dense support or opposition towards a political extremity \citep{chun2019detecting, tomaiuolo2020survey}.

\item Profile's Age: Assesses whether the account was created within a three-day window of one of the peaks of the discussion \citep{wang2023multi}.

\item Stock Market/Gambling (Betting) Content: Assesses whether an account frequently posts unrelated promotional content concerning stock markets, cryptocurrency, gambling, or betting services alongside political discourse-related hashtags. Such cross-topic promotional behavior has been associated with spam and bot accounts in prior research \citep{nizzoli2020charting}.
\end{enumerate}

\begin{table*}[ht!]
    \centering
    \caption{Number of user accounts processed during expert-annotation, and their distribution counts per criterion.}
    \begin{tabular}{lccccccccccc}
    \toprule
        Criteria & $N_{users}$ & 1 & 2 & 3 & 4 & 5 & 6 & 7 & 8 & 9 & 10\\\midrule
        24 hours & 102 & 96 & 96 & 96 & 96 & 97 & 95 & 96 & 96 & 98 & 98 \\
        tweet/retweet ratio & 101 & 96 & 96 & 96 & 96 & 96 & 96 & 96 & 96 & 96 & 96\\
        Additional & 115 & 112 & 112 & 112 & 112 & 112 & 112 & 112 & 112 & 112 & 112\\\bottomrule
    \end{tabular}
    \label{tab:annotation}
\end{table*}

Table~\ref{tab:annotation} shows the number of valid annotations obtained for each criterion in the rubric across the selected users. The smaller number of users for some criteria is due to factors such as account deletion or insufficient evidence. To ensure complete annotations across all criteria, we retained only users with valid annotations for every criterion, resulting in a labeled set of 301 users. This process required manually inspecting 154,032 posts, in addition to reviewing the users' profile information and tweets both within and outside the data collection period.

Additionally, annotators assigned one of three stance labels to the users based on their tweets about the proposed amendments under Law No. 7527. Pro-amendment users (222 accounts) were assigned label +1, anti-amendment users (78 accounts) were assigned -1, and only one user received a label 0 for having an undetermined stance. Since a single instance does not provide meaningful model training or evaluation, that user is removed from the annotated stance detection set, resulting in 300 users for stance modeling. Inter-annotator agreement for the stance labels was assessed using Cohen's kappa, returning near-perfect agreement ($\kappa = 0.94$).

\begin{figure*}[ht!]
    \centering

    \begin{subfigure}[b]{0.495\textwidth}
        \centering
        \includegraphics[width=\textwidth]{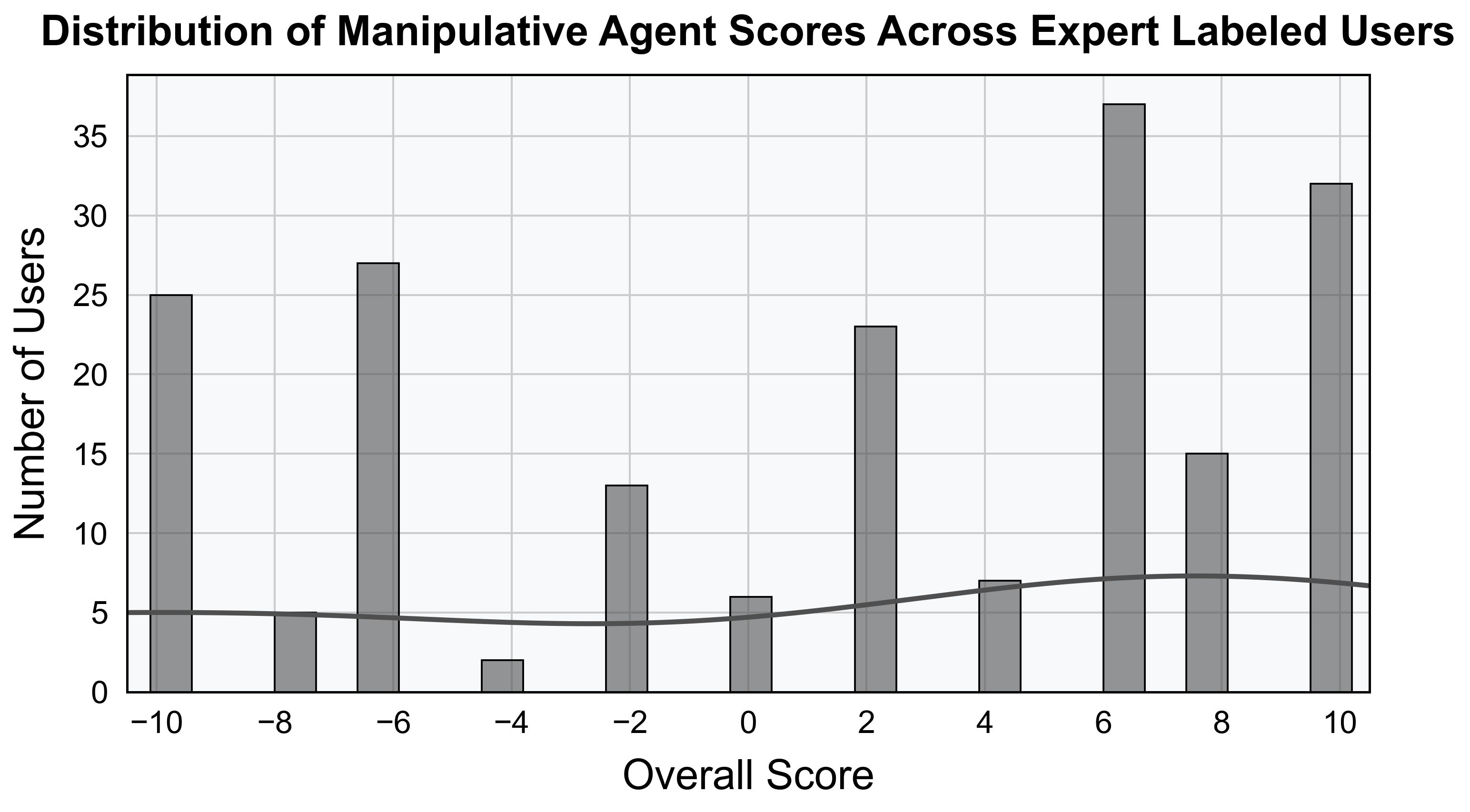}
        \caption{Distribution of manipulative participation scores.}
        \label{fig:expert_score_dist}
    \end{subfigure}
    \hfill
    \begin{subfigure}[b]{0.495\textwidth}
        \centering
        \includegraphics[width=\textwidth]{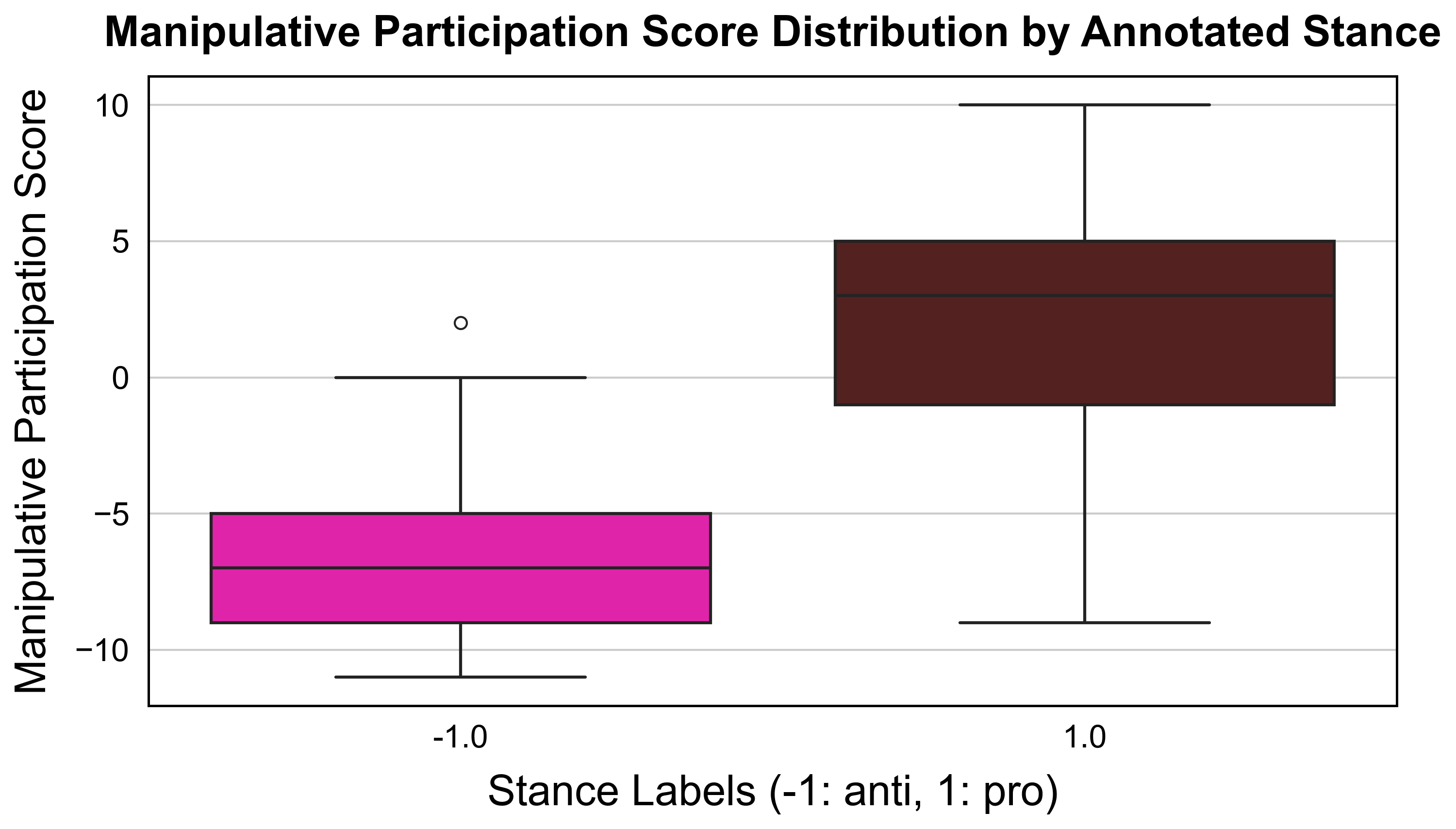}
        \caption{Scores and stance labels.}
        \label{fig:score_by_stance}
    \end{subfigure}

    \caption{Expert annotation distributions of the manipulative participation scores and stance labels as pro and anti-amendment.}
    \label{fig:expert_annotation_distributions}
\end{figure*}

The scores assigned across the ten criteria were summed to obtain each user's overall manipulative participation score. The resulting distribution is shown in Figure~\ref{fig:expert_score_dist}. All criteria were weighted equally because no empirical evidence justified differential weighting. Accordingly, negative summed scores indicate predominantly non-manipulative behavior, while higher positive values indicate stronger evidence of manipulative participation. The distribution of the 301 annotated users in Figure~\ref{fig:expert_score_dist} comprises 161 users with positive scores, 133 with negative scores, and 6 with a score of zero. Because manipulative participation was represented as a continuous score rather than categorical labels, Cohen's kappa is not appropriate for assessing labeler agreement. Therefore, we computed the intraclass correlation coefficient (ICC) for the continuous manipulative participation scores ranging from -10 to 10. We report both the single-rater and average-rater ICC values to quantify agreement between individual annotators and the reliability of the aggregated ratings used in subsequent analyses. 

The ICC for absolute agreement at the single-rater level is ICC(A,1) = 0.16 (95\% CI: -0.04, 0.35), while the ICC for the averaged ratings is ICC(C,k) = 0.44 (95\% CI: 0.30, 0.56). 
These values indicate weak-to-moderate agreement, consistent with the fact that manipulative participation is a latent behavioral construct inferred from multiple observable signals rather than directly measurable ground truth. 

After finalizing the expert-annotated dataset, we plot the relationship between the expert-annotated manipulative participation scores with the stance labels. Figure~\ref{fig:score_by_stance} shows that users with higher manipulative participation scores are more frequently associated with the pro-amendment stance (label +1), whereas lower or negative scores are more common among anti-amendment users (label -1). Considering how the annotators assigned scores to each ten criterion without considering the overall score, and also provided stance labels independently, this pattern suggests a potential relationship between manipulative participation and stance in the 2024 Street Animal Law debate. This relationship is further examined through prediction and generalization analyses in Section~\ref{sec:results}.

\subsection{Modeling framework}

\begin{figure*}
    \centering
    \includegraphics[width=0.9\textwidth]{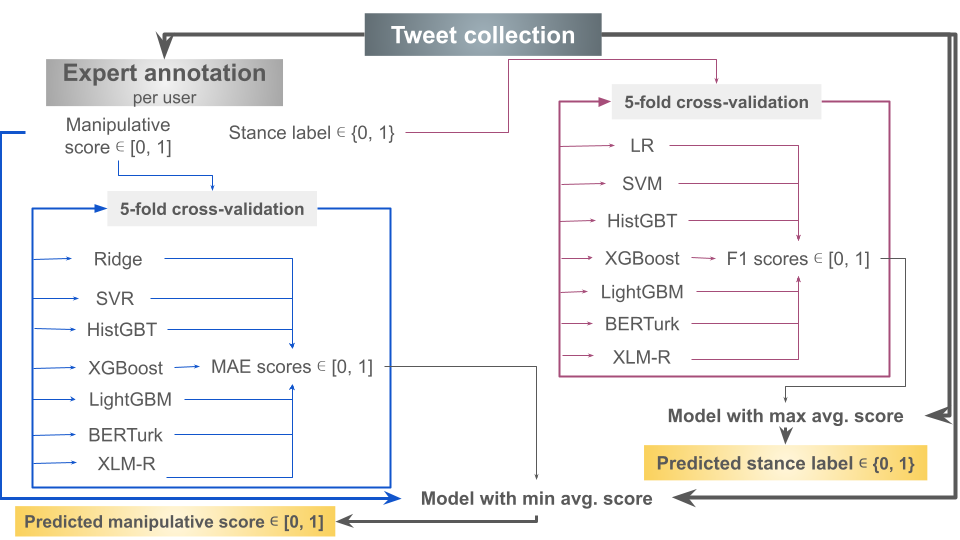}
    \caption{The unified framework developed for stance detection and manipulative participation filtering.}
    \label{fig:framework}
\end{figure*}

Figure~\ref{fig:framework} presents the proposed unified analytical framework. This framework estimates manipulative participation and user stance using a shared pipeline. Model evaluation within this framework is performed using user-level 5-fold cross-validation, described with more detail in Section~\ref{sec:evaluation}.

Because both stance labels and manipulative participation scores were assigned to sampled users based on their overall behavior, the unit of analysis throughout the framework is the user. However, because directly merging thousands of tweets into a single document is impractical for regression and classification, we assigned each user's labels and scores to all original tweets authored by that user during training. This corresponds to a label inheritance setting at the tweet level, where user-level annotations serve as proxy labels for individual tweets. We excluded all retweets of the users from the training process, assuming they primarily reflect content diffusion rather than original author expression. Predictions were generated at the tweet (and, in experiments incorporating retweets, at the retweet level) and subsequently aggregated to obtain user-level estimates. Manipulative participation scores were computed as the mean predicted score across a user's tweets, whereas stance was determined through majority voting over tweet-level predictions. The resulting user-level predictions are used throughout the subsequent analyses.

\subsection{Classical methods}

To establish baseline performance on both tasks, we evaluated a diverse set of classical machine learning models. These models rely on feature representations of the input text, making feature extraction an essential component of the pipeline.

\subsubsection{Feature representation}

For both tasks, we used the same feature extraction method that is standard in the NLP literature \citep{tomaiuolo2020survey, kuccuk2020stance}: word and word pair statistics, e.g., unigrams and bi-grams. Within each fold in the cross-validation, after lowercasing all posts, we extract n-gram ($n \in \{1, 2\}$) features that occur in at least five tweets within the dataset to eliminate typos and infrequent words. The extracted features are converted into TF-IDF (Term Frequency–Inverse Document Frequency) representations, with logarithmic term-frequency scaling applied. Feature selection is performed within each cross-validation fold using training data only to prevent information leakage.

Considering the high number of features present within the annotated set of thousands of tweets, feature selection is essential for scalable computation. Therefore, we applied a univariate ANOVA F-test on the training features to rank them according to their discriminative power and evaluated different numbers of top-ranked features $k \in \{250, 500, 750, 1000, 1500, 2000, 2500, 3000, 5000\}$, using Ridge regression for the manipulative participation scoring, and Logistic Regression for the stance detection. The optimal number of features, $k$, was selected via 5-fold cross-validating the training set by minimizing mean absolute error (MAE) for manipulative participation prediction and maximizing macro-F1 for stance classification. 

Previous studies have used user metadata such as their number of tweets, number of their followers, number of friends, number of likes, among others, for detecting bots using machine learning classifiers, and reported high performance \citep{abreu2020twitter}. Thus, we also evaluated the effect of including the meta data to the feature set. The user-level meta data available in our collection consisted of number of their followers and the number of users they follow. The tweet-level meta data we used are: number of views, retweets, quotes, favorites, replies, bookmarks. After extracting these metadata features, we applied standard scaling, concatenated them to the scaled TF-IDF features, and applied the same ANOVA feature selection.

\subsubsection{Manipulative participation estimation with regression}
To model manipulative participation as a continuous prediction task, we evaluated five regression algorithms with varying complexities: Ridge Regression, Support Vector Regression (SVR), Histogram-based Gradient Boosting Trees (HistGBT), Extreme Gradient Boosting (XGBoost), and Light Gradient Boosting Machine (LightGBM). Hyperparameters were optimized using Optuna with MAE minimized using 5-fold user-level cross-validation on the training set. The hyperparameter search spaces for these models are provided in Appendix~\ref{sec:optunaparams}.

\subsubsection{Stance detection}
We formulated the stance detection as a supervised binary classification problem after removing the single neutral case in the annotated set. Hyperparameter optimization was conducted using Optuna for all models and search spaces are provided in Appendix~\ref{sec:optunaparams}.

\subsection{Transformer models}

Unlike the classical machine learning models, transformers operate directly on tokenized text and therefore do not require manual feature engineering. We evaluated two pre-trained transformer encoders: BERTurk (dbmdz/bert-base-turkish-cased) \citep{stefan_schweter_2020_3770924}, following its strong performance in Turkish NLP applications \citep{sari2025vaccine}, and XLM-RoBERTa (XLM-R), a multilingual transformer encoder pretrained on CommonCrawl data across multiple languages \citep{xlmr}. XLM-R was included as a strong multilingual baseline to assess whether multilingual pretraining improves performance relative to a Turkish-specific model. For experiments incorporating metadata, we serialized that data as textual key-value pairs and prepended them to each tweet after tokenization, allowing the transformer to jointly encode textual content and metadata within a single input sequence. 

\begin{figure*}[ht!]
    \centering
    \includegraphics[width=\textwidth]{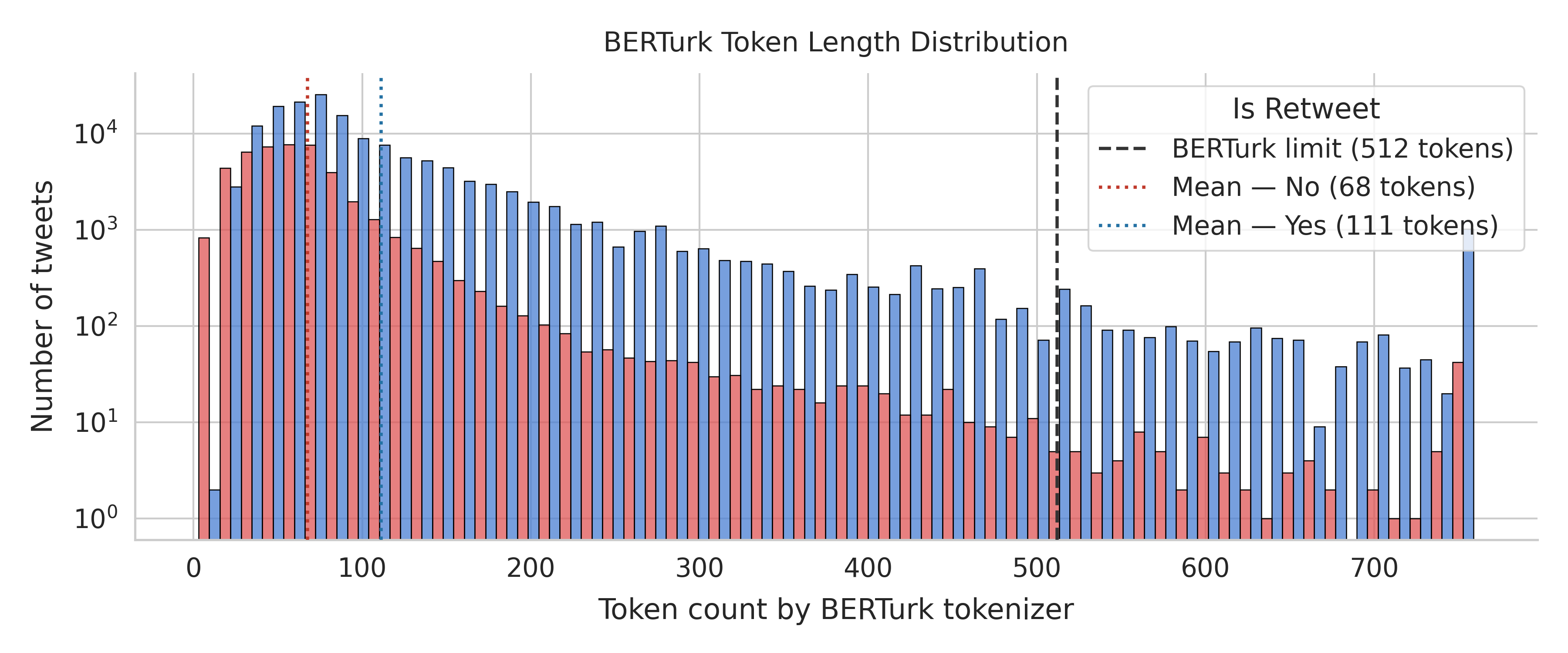}
    \caption{The distribution of token sizes of tweets and retweets in the dataset, together with their mean token sizes.}
    \label{fig:bertokens}
\end{figure*}

The hyperparameter search spaces used for optimizing the transformers for each task are provided in Appendix~\ref{sec:optunaparams}. 
 The AdamW optimizer was used with cross-entropy loss, and gradient norms were clipped at 1.0 for training stability. After tokenization with BERTurk, we analyzed the token length distribution (see Figure~\ref{fig:bertokens}) to select a maximum length. Considering this distribution while balancing computational cost and input coverage, we set this length to 150 tokens, preserving the majority of tokenized sequences. During the fine-tuning process, we reserved 20\% of the training set users as a user-level validation set to avoid information leakage and enable early stopping based on validation performance.

We used the same hyperparameter search ranges as above for the manipulative participation score prediction task. However, since this was a regression task, we performed a set of necessary alterations on the transformers. First, instead of a classification head with two neurons, we added a regression head with a single neuron. Second, we used the Mean Squared Error (MSE) loss, and during early stopping, we seeked the model with lowest MAE score on the validation set.

\subsection{Experimental design and evaluation}
\label{sec:evaluation}

We evaluated all models using 5-fold cross-validation on the expert-annotated user set. In each fold, 20\% of the annotated users were reserved for testing, while the remaining 80\% were used for training. Thus, each annotated user appeared in the test set exactly once. All splits were performed at the user level to ensure that no tweets from the same user were present in both the training and test partitions, preventing information leakage. Model performance is reported as the mean and standard deviation across the five folds. Finally, the best-performing model for each task was retrained on the complete annotated dataset and applied to the full collection to obtain manipulative participation scores and stance labels for all 537,994 users, enabling the subsequent computational social science analyses.

Regression performance was primarily evaluated using MAE, which directly measures the deviation between predicted and annotated manipulative participation scores. For completeness, we additionally report binary classification metrics obtained by thresholding the predicted manipulative participation scores at 0.5 in Appendix~\ref{ref:app}. Stance classification models were evaluated using macro-averaged precision, recall, and F1-score to account for class imbalance while giving equal importance to both stance classes. Retweets were excluded during model training but included when aggregating predictions for the complete collection.

All hyperparameter searches were conducted using 12 Optuna trials with pruning enabled. For the transformer experiments, we used a batch size of 64 and a maximum of 100 training epochs, although training always terminated earlier due to early stopping. All experiments were conducted on a workstation running Ubuntu 24.04.2 LTS equipped with an Intel Core i9-14900KF CPU (24 cores, 128 GB RAM) and two NVIDIA RTX 4500 Ada Generation GPUs (24 GB VRAM each). Models were implemented in Python 3.12.15 using Scikit-learn (1.6.1) \citep{pedregosa2011scikit}, Transformers (4.51.3), PyTorch (2.6.0+cu124), and Optuna (4.2.1) \citep{akiba2019optuna}.

\section{Results}

\subsection{Cross-validation}

First, we evaluate the performance of the selected methods on predicting continuous estimation of manipulative participation on the expert-annotated users within the 5-fold cross-validation framework. The resulting mean MAE values and standard deviations across five experiments are presented in Table~\ref{tab:troll_cv}. We repeated the same evaluation using identical user splits and random seeds with and without incorporating the metadata features. HistGBT and LightGBM achieved the lowest mean MAE values when only textual features were used, together with relatively low standard deviations across folds. Including metadata allowed Ridge regression to achieve a mean MAE comparable to the best text-only models, whereas the remaining machine learning methods generally showed no improvement and, in most cases, a slight decrease in performance. 

Both transformer models achieved MAE values comparable to those of the classical machine learning methods, but did not outperform the best-performing approaches and exhibited higher variability across folds, particularly BERTurk. Overall, several methods achieved similar predictive performance, with the lowest mean MAE values obtained by the gradient-boosting models using text-only features. HistGBT was selected for subsequent manipulative participation estimation.

\begin{table*}[ht!]
\centering
\caption{5-fold cross-validation mean and standard deviation of MAE for predicted continuous estimation of manipulative participation.}
\begin{tabular}{lrrrrrrr}
\toprule
Setting & Ridge & SVR & HistGBT & XGBoost & LightGBM & BERTurk & XLM-R \\\midrule
Text only & 0.2019  & 0.2061  & \textbf{0.1995}  & 0.2021  & \textbf{0.1984}  &  0.2002 &               0.2034 \\
 & $\pm$ 0.0099 & $\pm$ 0.0115 & $\pm$ \textbf{0.0117} & $\pm$ 0.0146 & $\pm$ \textbf{0.0134} & $\pm$ 0.0309 & $\pm$ 0.0144\\\hline
Text and metadata & \textbf{0.1990} & 0.2103  & 0.2207  & 0.2252  & 0.2284  & 0.2162 & 0.2312\\
& $\pm$ 0.0140  & $\pm$ 0.0160 & $\pm$ 0.0273 & $\pm$ 0.0264 & $\pm$ 0.0208 & $\pm$ 0.0252 & $\pm$ 0.0186 \\\bottomrule
\end{tabular}
\label{tab:troll_cv}
\end{table*}

Subsequent stance detection results are presented in Table~\ref{tab:stance_cv_results}. Unlike the estimation of manipulative participation, the best performance on this task was achieved by the transformer models, consistent with their ability to capture contextual information beyond the feature representations. While LR and LightGBM achieved the highest precision, BERTurk obtained the highest recall and macro-F1 score (0.8905) in text-only experiments, with consistently low variance across folds. XLM-R also performed competitively but remained below the Turkish-specific BERTurk model. These findings suggest that language-specific pretraining provides a measurable advantage for stance detection in Turkish social media data.

When metadata was included, BERTurk again achieved the highest macro-F1 score (0.8283) in Table~\ref{tab:stance_cv_results}, which is lower than the score achieved in the text-only experiment. Also, the metadata inclusion did not produce consistent improvements across the remaining models either. In several cases, performance slightly deteriorated. These results indicate that textual content alone captures most of the discriminative information required for both tasks. Consequently, the subsequent analyses rely on the text-only models, and metadata features are excluded from subsequent experiments. BERTurk was selected as the final stance classifier.

\begin{table*}[ht!]
\centering
\caption{5-fold cross-validation performance for user-level stance detection using textual features alone and textual features combined with metadata. Mean scores are reported in the first row and standard deviations in the second row.}
\begin{tabular}{lrrrrrr}
\toprule
& \multicolumn{3}{c}{Text only} & \multicolumn{3}{c}{Text and metadata} \\
\cmidrule(lr){2-4} \cmidrule(lr){5-7}
Model & Precision & Recall & F1 & Precision & Recall & F1 \\
\midrule
LR
& \textbf{0.9285} & 0.7920 & 0.8311
& \textbf{0.9213} & 0.7645 & 0.8031 \\
& $\pm$0.0379 & $\pm$0.0487 & $\pm$0.0481
& $\pm$0.0404 & $\pm$0.0612 & $\pm$0.0598 \\\hline

SVM
& 0.9140 & 0.7728 & 0.8064
& 0.9108 & 0.7380 & 0.7734 \\
& $\pm$0.0469 & $\pm$0.0769 & $\pm$0.0746
& $\pm$0.0535 & $\pm$0.0723 & $\pm$0.0782 \\\hline

HistGBT
& 0.9199 & 0.7721 & 0.8091
& 0.8766 & 0.7613 & 0.7910 \\
& $\pm$0.0513 & $\pm$0.0671 & $\pm$0.0730
& $\pm$0.0335 & $\pm$0.0522 & $\pm$0.0512 \\\hline

XGBoost
& 0.9093 & 0.7810 & 0.8153
& 0.8782 & 0.7526 & 0.7813 \\
& $\pm$0.0593 & $\pm$0.0718 & $\pm$0.0711
& $\pm$0.0466 & $\pm$0.0477 & $\pm$0.0465 \\\hline

LightGBM
& \textbf{0.9353} & 0.7867 & 0.8243
& 0.8678 & 0.7796 & 0.8037 \\
& $\pm$0.0272 & $\pm$0.0710 & $\pm$0.0658
& $\pm$0.0681 & $\pm$0.0377 & $\pm$0.0492 \\\hline

BERTurk
& \textbf{0.9293} & \textbf{0.8666} & \textbf{0.8905}
& 0.8723 & \textbf{0.8089} & \textbf{0.8283} \\
& $\pm$0.0332 & $\pm$0.0360 & $\pm$0.0262
& $\pm$0.0377 & $\pm$0.0487 & $\pm$0.0362 \\\hline

XLM-R
& 0.9234 & 0.8117 & 0.8432
& 0.8124 & 0.7673 & 0.7781 \\
& $\pm$0.0457 & $\pm$0.0781 & $\pm$0.0642
& $\pm$0.0798 & $\pm$0.0539 & $\pm$0.0599 \\
\bottomrule
\end{tabular}
\label{tab:stance_cv_results}
\end{table*}

\subsection{Generalization}

As the second stage of the unified framework in Figure~\ref{fig:framework}, the best-performing models identified during cross-validation, HistGBT for manipulative participation estimation and BERTurk for stance classification, were retrained using the complete annotated dataset and applied to the full collection. Hyperparameters and random seed settings remained identical to those used during cross-validation. For BERTurk, training-validation splitting is applied on the annotated-collection at the user-level with 80\%-20\% ratio for early stopping and parameter selection. During inference, predictions were generated for both original tweets and retweets before aggregating them to the user level. To assess robustness beyond cross-validation, and to further assess whether including retweets in applying models to the unseen data is useful, we independently expert-annotated an additional sample of 157 users drawn from the full collection, ensuring no overlap with the original annotated dataset. This sample served as an external validation set rather than as a development dataset.

\begin{table*}[ht!]
\centering
\caption{Generalization performance of the proposed unified framework on both tasks.}
\label{tab:gen}
\begin{tabular}{lll}
\toprule
Task & Retweets & Performance \\
\midrule
\multirow{2}{*}{Manipulative participation}
& Included & MAE = 0.2606 \\
& Excluded & MAE = 0.3073 \\
\midrule
\multirow{2}{*}{Stance classification}
& Included & Precision =  0.8555, Recall = 0.6911, F1 = 0.7234 \\
& Excluded & Precision = 0.8723, Recall = 0.6812, F1 =  0.7283\\
\bottomrule
\end{tabular}
\end{table*}

Generalization results on the independently annotated 157 users are presented in Table~\ref{tab:gen}. Both models maintained good performance on the external validation set, indicating that the proposed framework generalizes beyond the users used during model development. Including retweets improved manipulative participation estimation, reducing MAE from 0.3073 to 0.2606, whereas stance classification changed only marginally. These findings suggest that retweets provide complementary behavioral information for manipulative participation estimation but contribute little additional information for stance detection in our discourse analysis study.

\subsection{Post-analysis}

\begin{figure*}[ht!]
    \centering
    \begin{subfigure}[b]{0.495\textwidth}
    \includegraphics[width=\textwidth]{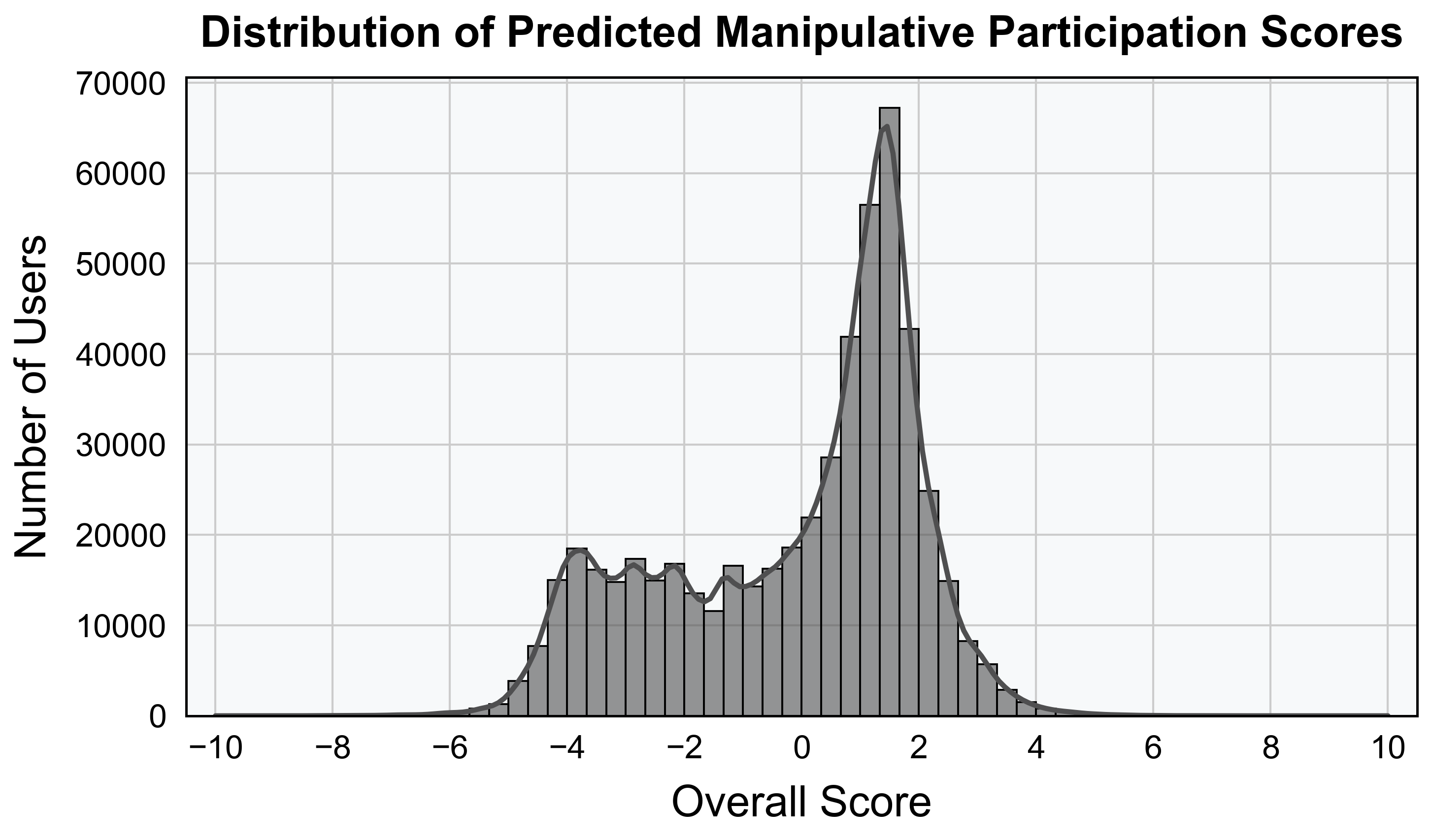}
    \caption{Distribution of the manipulative participation estimates}
    \label{fig:total_dist}
    \end{subfigure}
    \hfill
    \begin{subfigure}[b]{0.495\textwidth}
    \includegraphics[width=\textwidth]{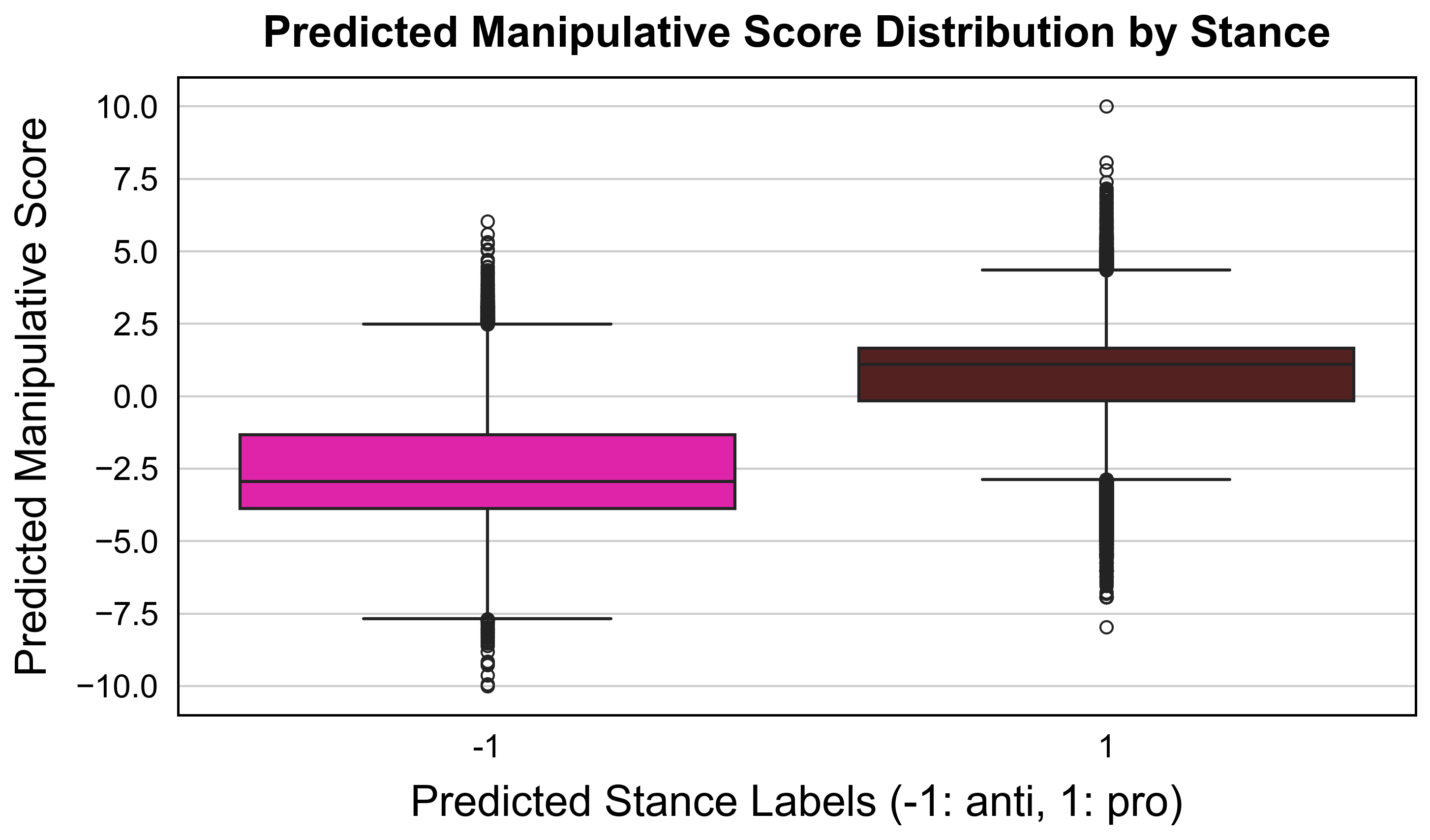}
    \caption{Manipulative participation estimates and stance labels}
    \label{fig:stance_vs_mani}
    \end{subfigure}

    \begin{subfigure}[b]{0.495\textwidth}
    \includegraphics[width=\textwidth]{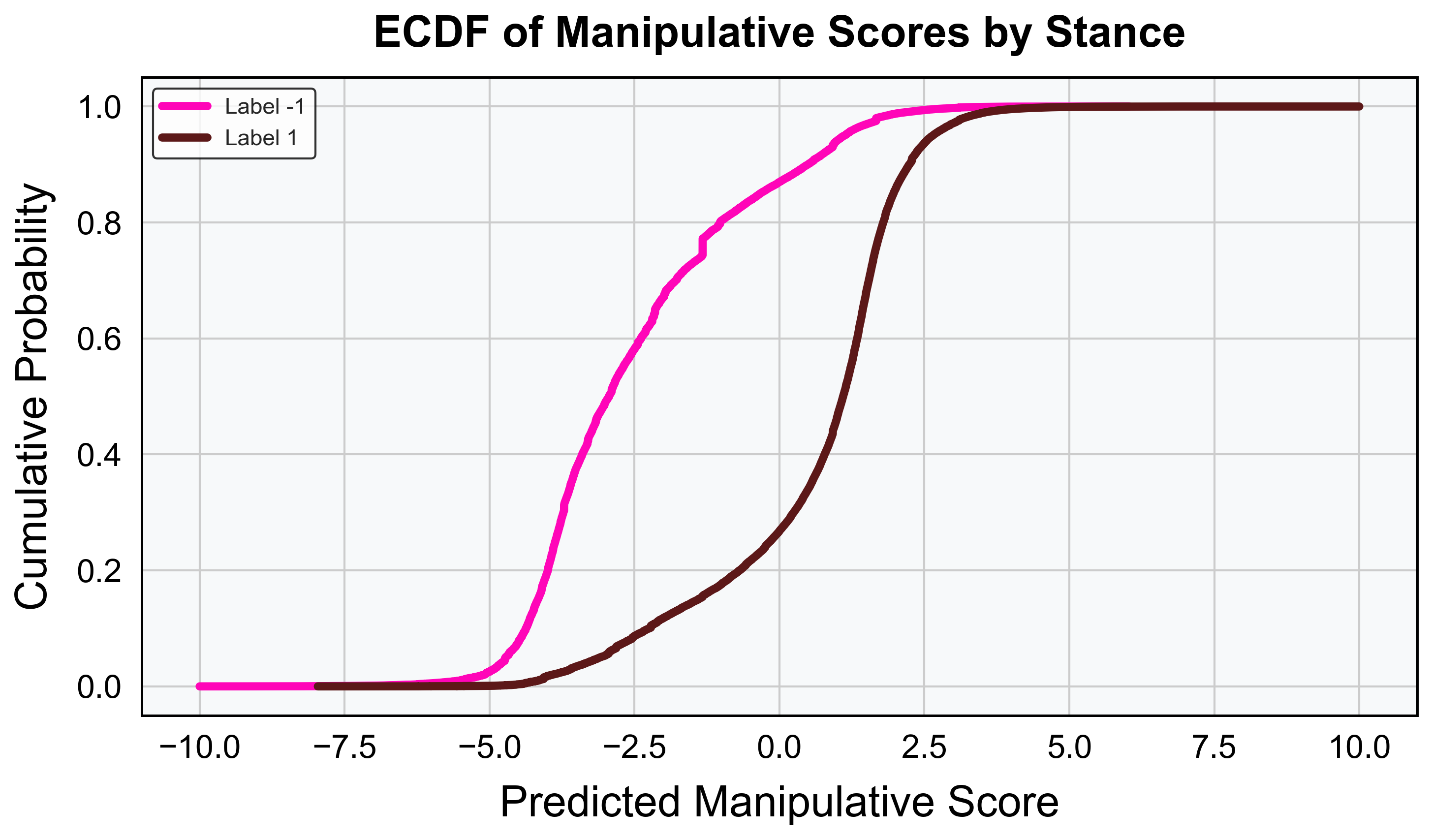}
    \caption{ECDF plot of the manipulative scores and stance labels}
    \label{fig:ecdf}
    \end{subfigure}
    \hfill
    \begin{subfigure}[b]{0.495\textwidth}
    \includegraphics[width=\textwidth]{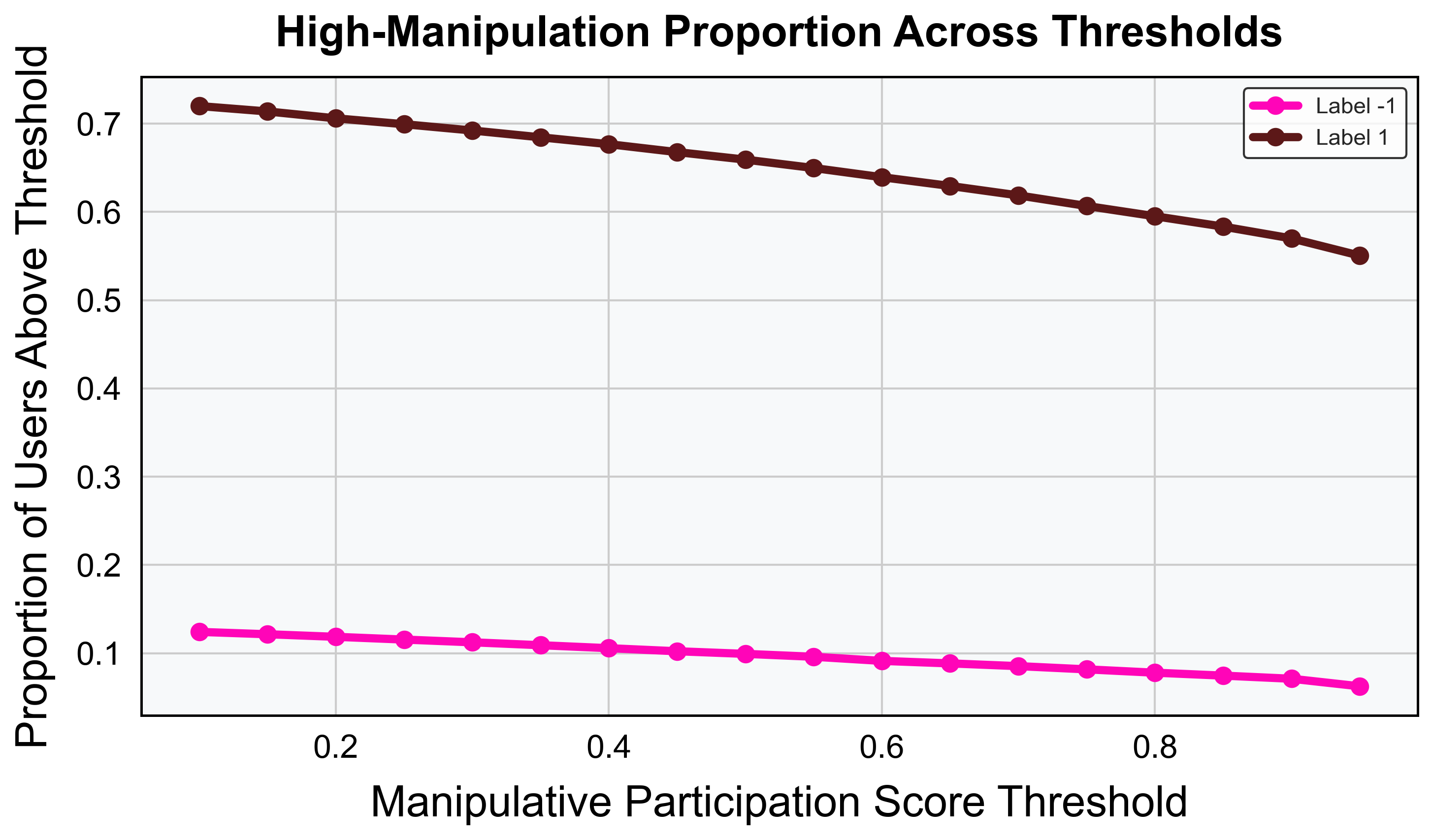}
    \caption{Different thresholds on manipulative participant population across stance labels}
    \label{fig:thresh}
    \end{subfigure}

    \begin{subfigure}[b]{0.7\textwidth}
    \includegraphics[width=\textwidth]{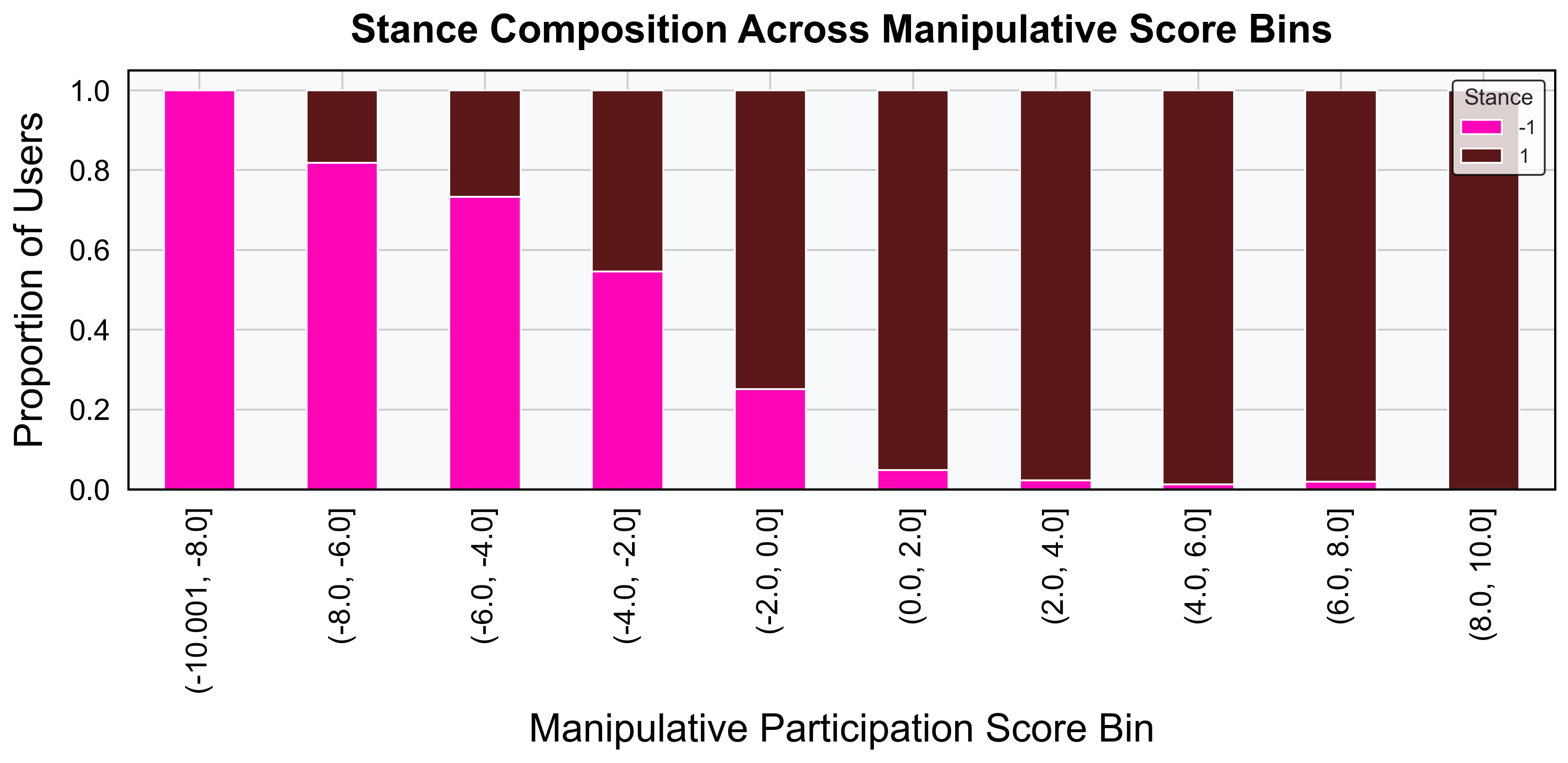}
    \caption{Stance labels across manipulative participation estimate bins}
    \label{fig:bins}
    \end{subfigure}

    \caption{Predicted results of the complete collection and their distributions per two tasks.}
    \label{fig:all_generalization}
\end{figure*}

Having established the generalization performance of the proposed framework, we next analyzed the predicted manipulative participation scores and stance labels across the complete collection. Figure~\ref{fig:total_dist} shows the distribution of estimated manipulative participation scores after rescaling them to the original $[-10,10]$ annotation range. The distribution is continuous rather than bimodal, with 59.27\% of users receiving positive scores. This supports the interpretation of manipulative participation as a graded behavioral characteristic rather than a binary attribute.

We also examined the relationship between the predicted manipulative participation scores and stance labels of the complete collection in Figures~\ref{fig:stance_vs_mani} to \ref{fig:bins}. Figure~\ref{fig:stance_vs_mani} shows that higher manipulative participation estimates are predominantly associated with the oppositional stance class, reproducing the trend observed in the manually annotated dataset (see Figure~\ref{fig:score_by_stance}). Figure~\ref{fig:ecdf} demonstrates that this separation persists across the entire score distribution rather than being confined to extreme values. Figure~\ref{fig:thresh} further shows that the association remains stable across a wide range of manipulative participation thresholds, indicating that it is not an artifact of selecting a particular cutoff. Finally, Figure~\ref{fig:bins} illustrates a monotonic increase in the proportion of oppositional users as manipulative participation estimates increase, providing additional evidence of a systematic relationship between the two predicted attributes. In the rare event of a tie between the two stance classes, the user was assigned an undetermined stance, and were excluded from these plots.

Collectively, these analyses demonstrate that treating manipulative participation as a continuous variable reveals systematic variation in stance that would be obscured by a binary manipulation versus genuine opinion distinction. This supports the use of continuous modeling when investigating the relationship between manipulative participation and political discourse.

To provide an additional qualitative assessment of the predicted stance labels, we conducted an additional manual inspection. We randomly sampled 100 users that were predicted as pro-amendment by the framework and submitted them for expert annotation. Of these, 70 were confirmed as pro-amendment, 21 were labeled as anti-amendment, and 9 remained undetermined. Although not intended as a formal evaluation, this manual inspection provided additional qualitative support for the predicted stance labels.

As we established the need for modeling manipulative participation as a continuous construct, we also explored the implications of this approach for downstream analyses. Specifically, computational social science studies that will use these predictions will need to convert them to binary classifications for tasks such as network analysis and descriptive group comparisons, which constitute part of our planned future work. Therefore, we continued the analysis by determining a threshold for filtering users based on their predicted manipulative participation scores using the expert-annotated dataset of 301 users. 

Data-driven thresholds derived from the predicted score distribution, such as the density antimode and Otsu's method, suggested threshold values of 0.33 and 0.385, respectively. However, these methods merely identify regions of low score density rather than an optimal classification boundary. Because these 301 users were used for model training, the resulting threshold should be interpreted as an operational decision boundary rather than an out-of-sample estimate. We therefore first converted the expert-annotated manipulative participation scores into a binary reference using 0.50 as the reference threshold and determined the optimal threshold on the model's predicted scores using the Youden index. This yielded an operational threshold of 0.55, which can be used to filter users classified as manipulative from the full collection for downstream analyses in our future work. We emphasize that this threshold is specific to the present dataset and should not be applied to other datasets without recalibration using the same approach.

\section{Discussion}

This study introduced a unified analytical framework for estimating manipulative participation and user stance from social media discourse, enabling analyses of how behavioral authenticity relates to observed opinion distributions. The results demonstrate that both tasks can be modeled within a common analytical pipeline while maintaining good predictive performance on annotated users. More importantly, applying the framework to the complete tweet collection of the Turkish street animals debate showed that manipulative participation and stance were not independent. Higher predicted manipulative participation scores were systematically associated with one stance in the debate, illustrating how manipulative activity may alter the apparent distribution of opinions observed on social media.

One of the principal methodological contributions of this study is representing manipulative participation as a continuous behavioral construct rather than a binary classification problem. Existing approaches frequently focus on identifying bots or coordinated accounts using binary labels, whereas the present results support representing manipulative participation as a continuous construct. Continuous scores preserve behavioral variation that is discarded by binary classification, enabling analyses of how increasing levels of manipulative participation relate to political stance. Rather than committing to a single decision boundary, researchers can examine the robustness of their findings across different manipulative participation thresholds or incorporate manipulative participation intensity directly into downstream analyses. The proposed framework therefore treats binary thresholding as a secondary calibration step. While the resulting threshold was 0.55 for the present dataset, future applications should estimate an operational decision boundary from their own annotated data instead of relying on a predefined cutoff such as 0.5.

The experiments also demonstrate that the two prediction tasks favor different model families. Gradient-boosting methods achieved the lowest prediction error for manipulative participation estimation, whereas transformer models, particularly BERTurk, produced the strongest stance classification performance. This difference is consistent with the supervision available for the two tasks. Stance labels correspond directly to the semantic content of tweets, allowing contextual language models to exploit linguistic information effectively. In contrast, manipulative participation is defined at the user level and inherited by individual tweets during training, creating a weakly supervised learning problem in which the prediction target reflects aggregate behavioral characteristics rather than the meaning of individual posts. Under these conditions, classical machine learning methods appear more robust than contextual transformer models. These findings demonstrate that higher-capacity models are not necessarily superior under weak supervision, emphasizing the importance of evaluating both classical machine learning and transformer-based approaches rather than assuming that a single state-of-the-art architecture will consistently provide the best performance.

The experiments further provide several practical observations for social media analysis in Turkish. First, BERTurk consistently outperformed the multilingual XLM-R model in stance classification, suggesting that language-specific pretraining better captures the linguistic characteristics of Turkish political discourse. Second, incorporating metadata did not consistently improve performance for either task, indicating that textual content alone contains most of the predictive information in the present dataset. Finally, including retweets during inference substantially improved manipulative participation estimation while producing only minor changes in stance detection. This suggests that retweeting behavior itself contains useful behavioral signals for estimating manipulative participation, whereas political stance is primarily expressed through users' original authored content.

Beyond prediction performance, the proposed framework has broader implications for computational social science and information science research. Social media discourse is frequently treated as an indicator of public opinion, yet online discussions may include varying degrees of manipulative participation that influence the visibility of particular viewpoints. By estimating stance and manipulative participation, the proposed framework allows researchers to evaluate how inferred opinion distributions change after accounting for varying levels of manipulative behavior rather than assuming that all observed activity reflects authentic user opinions. 

\section{Limitations and Future Work}

This study has several limitations. The proposed framework was evaluated on a single case study concerning public discourse surrounding a specific legislative process in Türkiye. Although this enabled detailed modeling of manipulative participation and user stance within a well-defined context, the annotation framework and behavioral criteria were developed for this setting. Applying the framework to different political topics, social media platforms, languages, or cultural contexts may require adapting both the annotation protocol and the sampling strategy.

Manipulative participation was annotated at the user level, while prediction models were trained at the tweet level by inheriting user-level labels. This weak supervision strategy enables efficient use of expert annotations but assumes that individual tweets represent the overall behavioral characteristics of their authors. In practice, individual posts may not always reflect the manipulative participation score assigned to the corresponding user. 
Similarly, quote-retweets may be used to criticize or ridicule the quoted content rather than endorse it. Because the quoted text typically constitutes most of the input sequence, the models may incorrectly infer the author's stance. Incorporating irony and sarcasm detection techniques may therefore further improve stance estimation. Although external validation demonstrated good generalization, stance classification has room for improvement, indicating that downstream analyses should be interpreted with appropriate caution and motivating further improvements in user-level stance estimation.

The proposed framework can be extended in several directions. Alternative supervision strategies, including tweet-level annotation, active learning, and semi-supervised learning, may reduce the potential limitations introduced by label inheritance. Second, evaluating the framework on different political topics, languages, and social media platforms would provide stronger evidence of its generalizability. Future work could also investigate more sophisticated approaches for user-level stance estimation, including probabilistic aggregation methods, confidence-aware prediction, and explicit modeling of users whose stance remains uncertain. Such approaches may provide a more nuanced representation than binary majority voting alone. 
Evaluating the framework across different political topics, languages, and social media platforms, together with improved user-level supervision strategies, would help establish the generalizability of continuous manipulative participation modeling for computational social science.

\section*{Acknowledgements}

This document is the results of the study supported by the Scientific and Technological Research Council of Türkiye (TÜBİTAK) under the 3005 Social Sciences and Humanities Research Support Program [Grant no 224K657].

\begin{appendices}
\section{Hyperparameters}
\label{sec:optunaparams}

For Ridge Regression, the regularization coefficient ($\alpha$) was searched on a logarithmic scale $\in [10^{-3}, 10^{2}]$. For SVR with an RBF kernel, the penalty parameter ($C$) was optimized $\in [10^{-2},10^{2}]$, the $\epsilon$-insensitive loss parameter $\in [10^{-3}, 1]$, and the kernel parameter $\gamma \in \{scale, auto\}$. For HistGBT, the learning rate was searched $\in [0.01,0.30]$, the number of boosting iterations between 50 and 300, the maximum tree depth between 2 and 10, and the L2 regularization term $\in [10^{-4}, 10]$. For XGBoost, the search space included the number of estimators $\in [50, 300]$, learning rate $\in [0.01, 0.30]$, and maximum depth $\in [2, 8]$. For LightGBM, the search space included the same range of number of estimators, and learning rate, with maximum depth $\in [2, 10]$, and number of leaves $\in [10, 100]$. All experiments were conducted using a fixed random seed to ensure reproducibility.

For the transformer models, hyperparameter optimization was performed separately for the stance detection and manipulative participation estimation tasks using Optuna. For stance detection, a classification head with two output neurons was added to each transformer. The search space included learning rate $\lambda \in [10^{-6},,5\times10^{-4}]$, weight decay $\in {0.0,0.2,0.4,0.6}$, dropout rate $\in {0.05,0.10,0.20,0.30,0.40}$, and early stopping patience $\in {3,4,5}$. For manipulative participation estimation, the same search space was used after replacing the classification head with a single-neuron regression head.

\section{Binary Evaluation of Manipulative Participation Prediction}
\label{ref:app}

For completeness, we additionally evaluated manipulative participation prediction after thresholding the predicted continuous scores into binary labels using the same 0.50 threshold adopted for the expert annotations. Precision, recall, and macro-F1 were then computed under the same 5-fold cross-validation protocol used throughout the study. These results are reported only as a supplementary comparison because manipulative participation is modeled as a continuous construct in the proposed framework.

\begin{table*}[ht!]
\centering
\caption{5-fold cross-validation mean and standard deviation of binary manipulative participation detection after thresholding regression predictions (text only).}
\begin{tabular}{lccc}
\hline
Model  & Precision & Recall & F1 \\
\hline
Ridge    &  0.7596 $\pm$ 0.0606 & 0.6812 $\pm$ 0.0362 & 0.6730 $\pm$ 0.0477 \\
SVR      &  0.7678 $\pm$ 0.0672 & 0.6892 $\pm$ 0.0367 & \textbf{0.6825} $\pm$ 0.0441 \\
HistGBT  &  \textbf{0.7740} $\pm$ 0.0616 & \textbf{0.6902} $\pm$ 0.0431 & \textbf{0.6831} $\pm$ 0.0549 \\
XGBoost  &  0.7334 $\pm$ 0.0683 & 0.6719 $\pm$ 0.0550 & 0.6654 $\pm$ 0.0657 \\
LightGBM &  0.7595 $\pm$ 0.0477 & 0.6851 $\pm$ 0.0291 & 0.6771 $\pm$ 0.0416 \\
BERTurk  &  0.6704 $\pm$ 0.2188 & 0.6533 $\pm$ 0.0924 & 0.6140 $\pm$ 0.1563 \\
XLM-R & 0.7563 $\pm$ 0.0524 & 0.6666 $\pm$ 0.0301 & 0.6515 $\pm$ 0.0364\\
\hline
\end{tabular}
\label{tab:troll_cv_app}
\end{table*}

\begin{table*}[ht!]
\centering
\caption{5-fold cross-validation mean and standard deviation of binary manipulative participation detection after thresholding regression predictions (text and metadata).}
\begin{tabular}{lccc}
\hline
Model & Precision & Recall & F1 \\
\hline
Ridge    & \textbf{0.7635} $\pm$ 0.0444 & \textbf{0.7015} $\pm$ 0.0473 & \textbf{0.6959} $\pm$ 0.0622 \\
SVR      & 0.6890 $\pm$ 0.0714 & 0.6534 $\pm$ 0.0726 & 0.6416 $\pm$ 0.0846 \\
HistGBT  & 0.6193 $\pm$ 0.0688 & 0.5961 $\pm$ 0.0608 & 0.5875 $\pm$ 0.0616 \\
XGBoost  & 0.6074 $\pm$ 0.0540 & 0.5966 $\pm$ 0.0467 & 0.5919 $\pm$ 0.0447 \\
LightGBM & 0.6142 $\pm$ 0.0361 & 0.6019 $\pm$ 0.0312 & 0.5979 $\pm$ 0.0307 \\
BERTurk  & 0.6240 $\pm$ 0.1898 & 0.6317 $\pm$ 0.0772 & 0.5955 $\pm$ 0.1384 \\
XLM-R & 0.5760 $\pm$ 0.1788 & 0.5946 $\pm$ 0.0572 & 0.5509 $\pm$ 0.1167\\
\hline
\end{tabular}
\label{tab:manipulation_cv_results_meta_app}
\end{table*}

Table~\ref{tab:troll_cv_app} and \ref{tab:manipulation_cv_results_meta_app} show that consistent with the regression results reported in the main text, HistGBT and SVR achieved the highest macro-F1 scores after thresholding the predicted manipulative participation scores, consistent with the regression results reported in the main text, HistGBT and SVR achieved the highest macro-F1 scores after thresholding the predicted manipulative participation scores. However, the differences among the strongest-performing models remained small, supporting the conclusions drawn from the continuous evaluation.

\end{appendices}

\bibliographystyle{unsrtnat}
\bibliography{references}  

@article{barberia2026sentiment,
  title={Sentiment and stance detection in COVID-19 vaccine discourse: a systematic review},
  author={Barberia, Lorena and Lombard, Belinda and Roman, Norton Trevisan and Sousa, Tatiane CM},
  journal={Journal of Computational Social Science},
  volume={9},
  number={3},
  pages={49},
  year={2026},
  publisher={Springer}
}

@article{reveilhac2022systematic,
  title={A systematic literature review of how and whether social media data can complement traditional survey data to study public opinion},
  author={Reveilhac, Maud and Steinmetz, Stephanie and Morselli, Davide},
  journal={Multimedia tools and applications},
  volume={81},
  number={7},
  pages={10107--10142},
  year={2022},
  publisher={Springer}
}

@article{schoch2022coordination,
  title={Coordination patterns reveal online political astroturfing across the world},
  author={Schoch, David and Keller, Franziska B and Stier, Sebastian and Yang, JungHwan},
  journal={Scientific reports},
  volume={12},
  number={1},
  pages={4572},
  year={2022},
  publisher={Nature Publishing Group UK London}
}

@article{sari2025vaccine,
  title={Vaccine hesitancy in T{\"u}rkiye: A natural language processing study on social media},
  author={Sari, Semih and Bayram, Ulya},
  journal={Turkish Journal of Electrical Engineering and Computer Sciences},
  volume={33},
  number={3},
  pages={392--409},
  year={2025},
  publisher={The Scientific and Technological Research Council of T{\"u}rkiye (T{\"U}B{\.I}TAK)}
}

@article{xlmr,
  author    = {Alexis Conneau and
               Kartikay Khandelwal and
               Naman Goyal and
               Vishrav Chaudhary and
               Guillaume Wenzek and
               Francisco Guzm{\'{a}}n and
               Edouard Grave and
               Myle Ott and
               Luke Zettlemoyer and
               Veselin Stoyanov},
  title     = {Unsupervised Cross-lingual Representation Learning at Scale},
  journal   = {CoRR},
  volume    = {abs/1911.02116},
  year      = {2019},
  url       = {http://arxiv.org/abs/1911.02116},
  eprinttype = {arXiv},
  eprint    = {1911.02116},
  bibsource = {dblp computer science bibliography, https://dblp.org}
}

@inproceedings{elmas2023analyzing,
  title={Analyzing activity and suspension patterns of twitter bots attacking turkish twitter trends by a longitudinal dataset},
  author={Elmas, Tu{\u{g}}rulcan},
  booktitle={Companion Proceedings of the ACM Web Conference 2023},
  pages={1404--1412},
  year={2023}
}

@inproceedings{samih2021few,
  title={A few topical tweets are enough for effective user stance detection},
  author={Samih, Younes and Darwish, Kareem},
  booktitle={Proceedings of the 16th Conference of the European Chapter of the Association for Computational Linguistics: Main Volume},
  pages={2637--2646},
  year={2021}
}

@inproceedings{kreutz2022detecting,
  title={Detecting vaccine skepticism on twitter using heterogeneous information networks},
  author={Kreutz, Tim and Daelemans, Walter},
  booktitle={International Conference on Applications of Natural Language to Information Systems},
  pages={370--381},
  year={2022},
  organization={Springer}
}

@article{zhao2026leveraging,
  title={Leveraging deep active learning for multimodal health misinformation detection on social media},
  author={Zhao, Yuehua and Zeng, Shiyang},
  journal={Information Processing \& Management},
  volume={63},
  number={3},
  pages={104540},
  year={2026},
  publisher={Elsevier}
}

@article{aldayel2022characterizing,
  title={Characterizing the role of bots’ in polarized stance on social media},
  author={Aldayel, Abeer and Magdy, Walid},
  journal={Social Network Analysis and Mining},
  volume={12},
  number={1},
  pages={30},
  year={2022},
  publisher={Springer}
}

@article{cinelli2022coordinated,
  title={Coordinated inauthentic behavior and information spreading on Twitter},
  author={Cinelli, Matteo and Cresci, Stefano and Quattrociocchi, Walter and Tesconi, Maurizio and Zola, Paola},
  journal={Decision Support Systems},
  volume={160},
  pages={113819},
  year={2022},
  publisher={Elsevier}
}

@article{nizzoli2020charting,
  title={Charting the landscape of online cryptocurrency manipulation},
  author={Nizzoli, Leonardo and Tardelli, Serena and Avvenuti, Marco and Cresci, Stefano and Tesconi, Maurizio and Ferrara, Emilio},
  journal={IEEE access},
  volume={8},
  pages={113230--113245},
  year={2020},
  publisher={IEEE}
}

@article{antonakaki2026detecting,
  title={Detecting Synthetic Political Narratives in Cross-Platform Social Media Discourse},
  author={Antonakaki, Despoina and Ioannidis, Sotiris},
  journal={arXiv preprint arXiv:2605.21540},
  year={2026}
}

@article{rauchfleisch2020false,
  title={The false positive problem of automatic bot detection in social science research},
  author={Rauchfleisch, Adrian and Kaiser, Jonas},
  journal={PloS one},
  volume={15},
  number={10},
  pages={e0241045},
  year={2020},
  publisher={Public Library of Science San Francisco, CA USA}
}

@online{GazeteOksijen2024,
  author       = {{Gazete Oksijen}},
  title        = {Konda anketi: Yüzde 85 sokak hayvanlarının uyutulmasına karşı},
  year         = {2024},
  month        = jul,
  day          = {20},
  url          = {https://gazeteoksijen.com/turkiye/konda-anketi-yuzde-85-sokak-hayvanlarinin-uyutulmasina-karsi-217412},
  note         = {Accessed: 2026-05-28}
}

@article{kuccuk2020stance,
  title={Stance detection: A survey},
  author={K{\"u}{\c{c}}{\"u}k, Dilek and Can, Fazli},
  journal={ACM Computing Surveys (CSUR)},
  volume={53},
  number={1},
  pages={1--37},
  year={2020},
  publisher={ACM New York, NY, USA}
}

@online{GercekGundem2024,
  author       = {{Gerçek Gündem}},
  title        = {Ezici çoğunluk aynı cevabı verdi: Türkiye’nin en önemli sorunu ne? [The overwhelming majority gave the same answer: What is Turkey's most important problem?]},
  year         = {2024},
  month        = jul,
  day          = {27},
  url          = {https://www.gercekgundem.com/ekonomi/ezici-cogunluk-ayni-cevabi-verdi-turkiyenin-en-onemli-sorunu-ne-472525},
  note         = {Accessed: 2026-05-28}
}

@article{alav2025turkish,
  title={Turkish stance detection on social media using bert models: A case study of stray animals law},
  author={Alav, Selma and Benli, Kristin Surpuhi},
  journal={Sakarya University Journal of Computer and Information Sciences},
  volume={8},
  number={1},
  pages={76--88},
  year={2025},
  publisher={Sakarya University}
}

@inproceedings{abreu2020twitter,
  title={Twitter bot detection with reduced feature set},
  author={Abreu, Jefferson Viana Fonseca and Ralha, C{\'e}lia Ghedini and Gondim, Jo{\~a}o Jos{\'e} Costa},
  booktitle={2020 IEEE International Conference on Intelligence and Security Informatics (ISI)},
  pages={1--6},
  year={2020},
  organization={IEEE}
}

@article{rodriguez2020one,
  title={A one-class classification approach for bot detection on Twitter},
  author={Rodr{\'\i}guez-Ruiz, Jorge and Mata-S{\'a}nchez, Javier Israel and Monroy, Raul and Loyola-Gonz{\'a}lez, Octavio and L{\'o}pez-Cuevas, Armando},
  journal={Computers \& Security},
  volume={91},
  pages={101715},
  year={2020},
  publisher={Elsevier}
}

@article{wu2019misinformation,
  title={Misinformation in social media: definition, manipulation, and detection},
  author={Wu, Liang and Morstatter, Fred and Carley, Kathleen M and Liu, Huan},
  journal={ACM SIGKDD explorations newsletter},
  volume={21},
  number={2},
  pages={80--90},
  year={2019},
  publisher={ACM New York, NY, USA}
}

@article{molenda2022makes,
  title={What makes an Internet troll? On the relationships between temperament (BIS/BAS), Dark Triad, and Internet trolling},
  author={Molenda, Zuzanna Aleksandra and Marchlewska, Marta and Rogoza, Marta and Michalski, Piotr and G{\'o}rska, Paulina and Szczepa{\'n}ska, Dagmara and Cislak, Aleksandra},
  journal={Cyberpsychology: Journal of Psychosocial Research on Cyberspace},
  volume={16},
  number={5},
  year={2022}
}

@misc{stefan_schweter_2020_3770924,
  author       = {Stefan Schweter},
  title        = {BERTurk - BERT models for Turkish},
  month        = apr,
  year         = 2020,
  publisher    = {Zenodo},
  version      = {1.0.0},
  doi          = {10.5281/zenodo.3770924},
  url          = {https://doi.org/10.5281/zenodo.3770924}
}

@inproceedings{akiba2019optuna,
  title={Optuna: A next-generation hyperparameter optimization framework},
  author={Akiba, Takuya and Sano, Shotaro and Yanase, Toshihiko and Ohta, Takeru and Koyama, Masanori},
  booktitle={Proceedings of the 25th ACM SIGKDD international conference on knowledge discovery \& data mining},
  pages={2623--2631},
  year={2019},
  doi={10.1145/3292500.3330701}
}

@article{pedregosa2011scikit,
  title={Scikit-learn: Machine learning in python journal of machine learning research},
  author={Pedregosa, Fabian and Varoquaux, Gael and Gramfort, Alexandre and Michel, Vincent and Thirion, Bertrand and Grisel, Olivier and Blondel, Mathieu and Prettenhofer, Peter and Weiss, Ron and Dubourg, Vincent and others},
  journal={Journal of machine learning research},
  volume={12},
  pages={2825--2830},
  year={2011},
  doi={10.5555/1953048.2078195}
}

@article{yang2022botometer,
  title={Botometer 101: Social bot practicum for computational social scientists},
  author={Yang, Kai-Cheng and Ferrara, Emilio and Menczer, Filippo},
  journal={Journal of computational social science},
  volume={5},
  number={2},
  pages={1511--1528},
  year={2022},
  publisher={Springer}
}

@article{suler2004online,
  title={The online disinhibition effect},
  author={Suler, John},
  journal={Cyberpsychology \& behavior},
  volume={7},
  number={3},
  pages={321--326},
  year={2004},
  publisher={Mary Ann Liebert, Inc.}
}

@article{coles2016trolling,
  title={Trolling the trolls: Online forum users constructions of the nature and properties of trolling},
  author={Coles, Bryn Alexander and West, Melanie},
  journal={Computers in Human Behavior},
  volume={60},
  pages={233--244},
  year={2016},
  publisher={Elsevier}
}

@article{tomaiuolo2020survey,
  title={A survey on troll detection},
  author={Tomaiuolo, Michele and Lombardo, Gianfranco and Mordonini, Monica and Cagnoni, Stefano and Poggi, Agostino},
  journal={Future internet},
  volume={12},
  number={2},
  pages={31},
  year={2020},
  publisher={MDPI}
}

@inproceedings{chun2019detecting,
  title={Detecting Political Bias Trolls in Twitter Data.},
  author={Chun, Soon Ae and Holowczak, Richard D and Dharan, Kannan and Wang, Ruoyu and Basu, Soumaydeep and Geller, James},
  booktitle={WEBIST},
  pages={334--342},
  year={2019}
}

@inproceedings{monakhov2020understanding,
  title={Understanding troll writing as a linguistic phenomenon},
  author={Monakhov, Sergei},
  booktitle={Proceedings of SAI Intelligent Systems Conference},
  pages={315--334},
  year={2020},
  organization={Springer}
}

@article{fornacciari2018holistic,
  title={A holistic system for troll detection on Twitter},
  author={Fornacciari, Paolo and Mordonini, Monica and Poggi, Agostino and Sani, Laura and Tomaiuolo, Michele},
  journal={Computers in Human Behavior},
  volume={89},
  pages={258--268},
  year={2018},
  publisher={Elsevier}
}

@inproceedings{seah2015troll,
  title={Troll detection by domain-adapting sentiment analysis},
  author={Seah, Chun Wei and Chieu, Hai Leong and Chai, Kian Ming A and Teow, Loo-Nin and Yeong, Lee Wei},
  booktitle={2015 18th International Conference on Information Fusion (Fusion)},
  pages={792--799},
  year={2015},
  organization={IEEE}
}

@inproceedings{wang2023multi,
  title={A Multi-Modal Approach for the Detection of Account Anonymity on Social Media Platforms},
  author={Wang, Bo and Guo, Jie and Huang, Zheng and Qiu, Weidong},
  booktitle={2023 International Joint Conference on Neural Networks (IJCNN)},
  pages={1--7},
  year={2023},
  organization={IEEE}
}

\end{document}